\documentclass[letterpaper]{article} 
\usepackage[preprint]{EviDet}  
\usepackage[hyphens]{url}  
\usepackage{graphicx} 
\usepackage{natbib}  
\usepackage{caption} 
\usepackage{algorithm}
\usepackage{algorithmic}
\usepackage{amsmath} 
\usepackage{amssymb} 
\usepackage{bm}

\usepackage{newfloat}
\usepackage{listings}
\DeclareCaptionStyle{ruled}{labelfont=normalfont,labelsep=colon,strut=off} 
\floatstyle{ruled}
\newfloat{listing}{tb}{lst}{}
\floatname{listing}{Listing}

\usepackage{booktabs}
\usepackage{multirow}

\title{EviDet: Learning Decision-Relevant Evidence for AI-Generated Image Detection and Faithful Explanation}
\author{
    Jinchun Zhang\textsuperscript{\rm 1}\equalcontrib,
    Hao Chen\textsuperscript{\rm 1}\equalcontrib,
    Peipeng Yu\textsuperscript{\rm 2},
    Zhihua Xia\textsuperscript{\rm 1}\thanks{Corresponding author: xia\_zhihua@163.com.},
    Xiaoyu Zhou\textsuperscript{\rm 1}
}
\affiliations{
    \textsuperscript{\rm 1}College of Cyber Security, Jinan University\\
    \textsuperscript{\rm 2}School of Cyber Science and Engineering, Nanjing University of Information Science and Technology\\
}

\begin{document}

\maketitle

\begin{abstract}
As generative models make AI-generated images harder to distinguish from real ones, reliable image forensics is essential to identify misinformation. Recent methods combine detection with natural-language explanations, yet rarely test whether these explanations reflect the evidence behind the decision. Such explanations may read as plausible but not faithful, describing artifacts absent from the image. To address this gap, we introduce \textbf{EviDet}, a framework that learns \textbf{a decision-relevant evidence map ($E$)} to anchor both detection and explanation. In the detection stage, low-level artifacts are extracted as candidate cues by the Multi-Scale Resampling Prior (\textbf{MSRP}) module and fused with self-supervised features to form $E$, guiding fine-grained classification over real, fully synthetic, and tampered images. The Evidence Masking Verification (\textbf{EMV}) module then intervenes on features during training, driving $E$ to reflect the decision's evidence. In the explanation stage, an Evidence-Guided Explainer (\textbf{EGE}) reduces unsupported forensic claims by grounding an LLM in $E$ via evidence-null contrastive decoding. Extensive experiments demonstrate state-of-the-art performance: 99.4\% accuracy on SID-Set, 97.7\% accuracy on unseen diffusion generators of OpenSDID, and a reduction of the hallucination rate (CHAIR) from 0.73 to 0.54 over standard decoding. These results support the evidence map $E$ as a measurable link between accurate detection and faithful explanation.
\end{abstract}


\section{Introduction}

Rapid advances in generative models have made AI-generated images (AIGIs) increasingly realistic, blurring the boundary between authentic and synthetic content. Forged images generally take two forms: fully synthetic images are generated end-to-end, whereas tampered images involve replacing or modifying specific objects or local regions through generative editing~\cite{huang2025sida}. The latter can leverage the credibility of real-world scenes, making targeted edits more difficult to detect. Misuse of either form threatens information integrity and public trust. Addressing these threats calls for reliable AIGI detection methods that provide both accurate predictions and faithful explanations grounded in the evidence supporting them.

\begin{figure}[t]
\centering
\includegraphics[width=1.0\columnwidth]{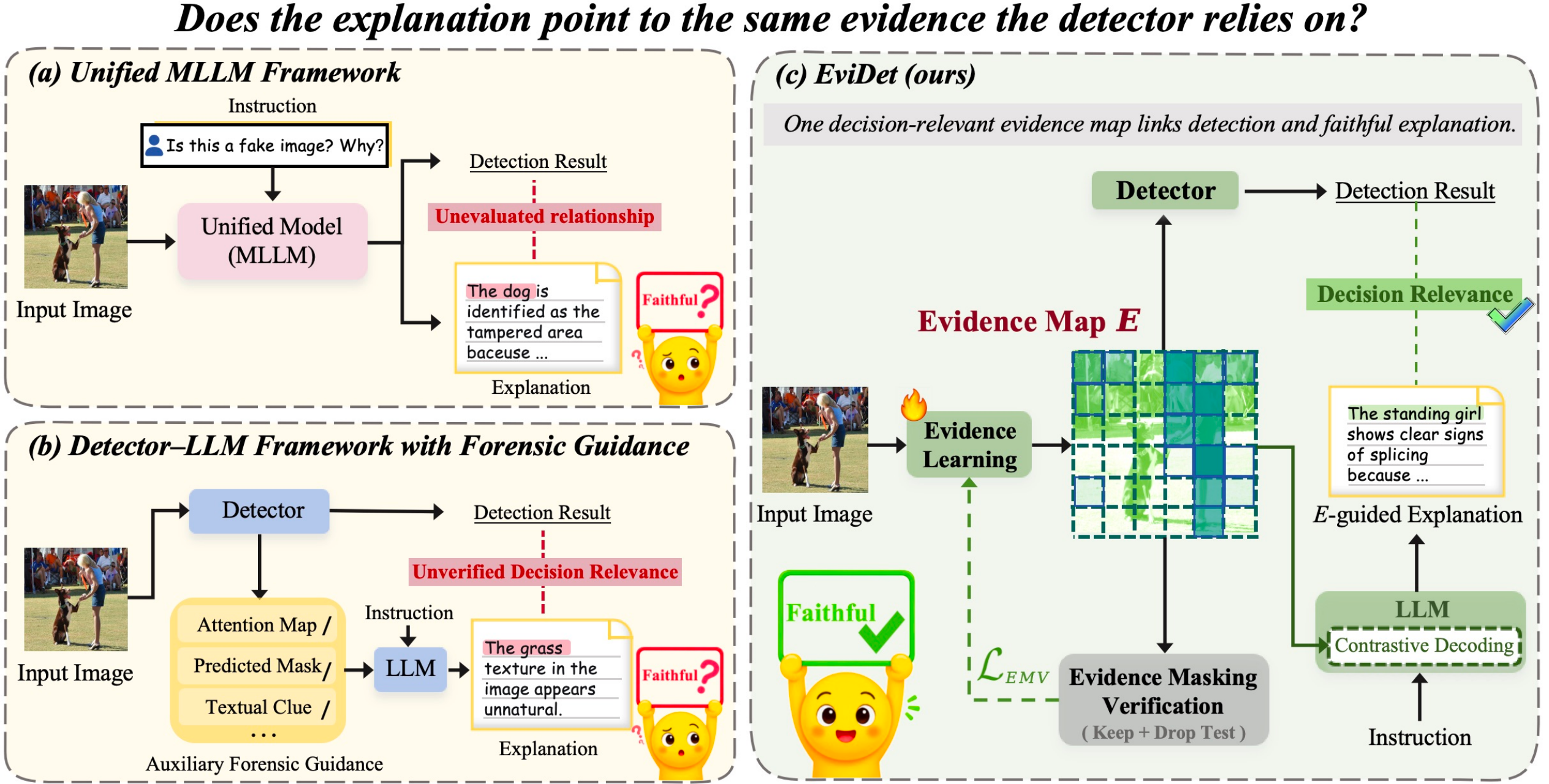} 
\caption{Existing methods do not directly test whether explanations refer to evidence that influences the detector’s decision. (a) Unified MLLM frameworks produce the classification result and the explanation without checking their consistency. (b) Detector--LLM frameworks pass auxiliary cues to an LLM without verifying decision relevance. (c) EviDet (ours) learns a patch-level evidence map ($E$), checks its decision relevance via feature-level interventions, and conditions the explanation on it.}
\label{fig:teaser}
\end{figure}

Research on AIGI detection has primarily targeted cross-generator generalization, learning from low-level forensic artifacts~\cite{tan2024npr,wang2023dire,ricker2024aeroblade,karageorgiou2025any}, adapting large pretrained vision models~\cite{tan2025c2pclip,chen2025forgelens,ojha2023univfd}, or fusing forensic and semantic features~\cite{yan2025aide,cheng2025cospy}. While these approaches demonstrate that certain artifacts transfer across generators, they fail to establish whether a detector relies on them for any given image. Label-only supervision constrains the output label, not the basis of the decision, allowing detectors to exploit spurious correlations with spatial position, semantic content, or file format ~\cite{fu2025udd,guillaro2025bias,cai2025cataid}. Such correlations may fail under generator or content shifts, thereby undermining generalization. This motivates a distinction between an \emph{artifact} (an observable trace that may indicate manipulation) and \emph{evidence} (an artifact shown to support the detector's prediction). The same gap can further affect explanations, which may rationalize a prediction post hoc rather than faithfully reflect the evidence used by the detector.

More recently, multimodal large language models (MLLMs) have extended detection from a black-box score to a human-readable analysis. As illustrated in Figure~\ref{fig:teaser} (a)--(b), existing approaches either employ a unified MLLM that jointly produces the verdict and explanation~\cite{huang2025sida,xu2025fakeshield,kang2025legion,wen2025fakevlm,ji2026fakexplain}, or adopt a two-stage pipeline that conditions a large language model (LLM) on detection-stage auxiliary cues, such as attention maps, predicted masks, or textual forensic cues~\cite{guo2025m2f2,yu2025unlocking,zhou2025aigi}. However, neither design explicitly verifies the link between a prediction and its explanation. Fine-tuning on reference explanations shapes the phrasing of forensic accounts but not their fidelity to the detector's evidence, while two-stage pipelines pass auxiliary cues to the LLM without checking that those cues support the prediction or that the LLM actually relies on them. Consistent with this gap, recent evaluations find such explanations often lack visual grounding and assert unverifiable artifacts~\cite{jiang2026tridf,kuckreja2026deepfakejudge}. Such explanations can be \emph{plausible} but not \emph{faithful}: an explanation may cite artifacts absent from the image, which we call a \emph{forensic hallucination}, or cues present in the image that played no part in the decision. In high-stakes settings such as judicial review and journalistic fact-checking, a wrong verdict backed by a convincing but unsupported explanation is harder to catch and more damaging than a bare misclassification. This concern is amplified in the fine-grained setting of SIDA~\cite{huang2025sida}, where only tampered images come with ground-truth masks while fully synthetic ones provide no spatial supervision at all, leaving spatial grounding asymmetric across classes.

Overall, existing methods lack a representation that is simultaneously used by the classifier, checked for predictive relevance, and used to ground the explanation. To address this gap, we introduce \textbf{EviDet}, a framework that learns decision-relevant evidence for AI-generated image detection and faithful explanation following the fine-grained setting introduced by SIDA~\cite{huang2025sida}, which spans real, fully synthetic, and tampered images. As illustrated in Figure~\ref{fig:teaser}(c), EviDet closes this loop with a patch-level evidence map $E$ that plays all three roles. 

EviDet rests on a strict distinction between \emph{artifacts} and \emph{evidence}. Specifically, the Multi-Scale Resampling Prior (\textbf{MSRP}) module extracts candidate forensic artifacts from residual responses induced by downscaling and restoring the input at multiple scales. Fused with adapted DINOv3
features~\cite{simeoni2025dinov3} via cross-attention, they initialize $E$, which weights the pooled features feeding the classifier and is learned jointly under classification and class-specific supervision. During training, the Evidence Masking Verification (\textbf{EMV}) module intervenes at the \textbf{feature level} by retaining or suppressing the patches $E$ selects, which drives $E$ to reflect the \textbf{decision relevance} of the evidence: keeping only these patches must preserve the fake prediction, and removing them must lower it. Once the detector is trained and frozen, the Evidence-Guided Explainer (\textbf{EGE}) conditions a language model on the detector features, the predicted class, and $E$, contrasting evidence-present against evidence-null decoding to amplify tokens that the evidence supports. A shared decision-relevant evidence map ($E$) thus connects detection and explanation.

Our main contributions are summarized as follows:
\begin{itemize}
    \item We reframe AIGI detection around the distinction between \emph{artifacts} and \emph{evidence}: \textbf{MSRP} extracts generalizable low-level artifacts as candidate cues, distilled into a patch-level evidence map $E$ that \textbf{EMV} checks for decision relevance and that grounds the explanation.
    \item We propose \textbf{EGE}, which anchors an LLM to the evidence map through evidence-null contrastive training and decoding. This yields explanations grounded in the detector's evidence map with fewer unsupported claims.
    \item Extensive experiments demonstrate that EviDet achieves state-of-the-art performance: 99.4\% three-class accuracy on SID-Set, 97.7\% binary accuracy on unseen OpenSDID generators, and reduces the hallucination rate (CHAIR) from 0.73 to 0.54 on TriDF~\cite{jiang2026tridf}.
\end{itemize}

\begin{figure*}[t]
\centering
\includegraphics[width=0.9\textwidth]{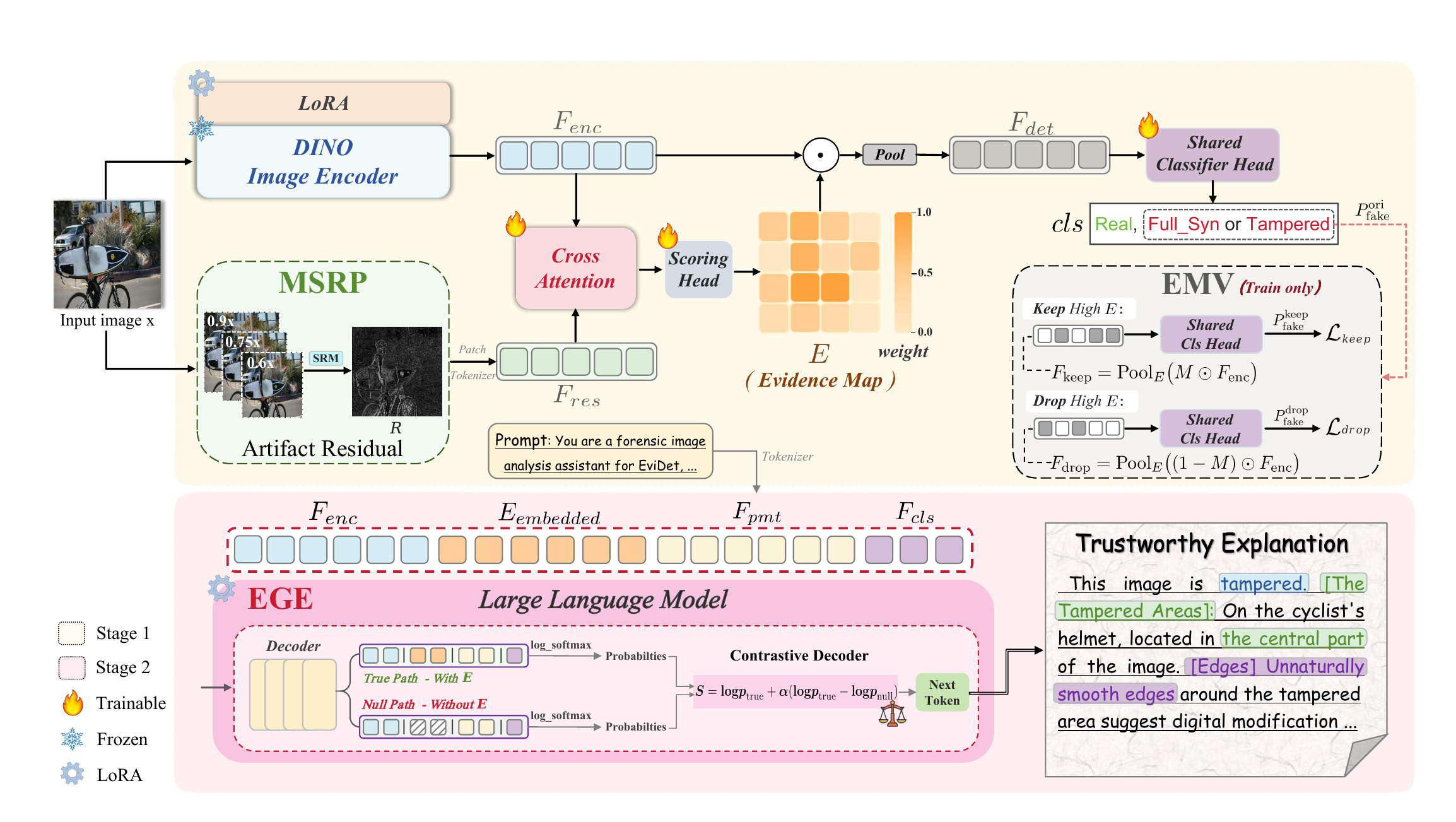} 
\caption{Overview of EviDet. In Stage 1, the Multi-Scale Resampling Prior (\textbf{MSRP}) extracts low-level artifacts that are fused with LoRA-adapted DINOv3 features and scored into the evidence map $E$, gating the three-way classifier. Evidence Masking Verification(\textbf{EMV}) then applies keep and drop interventions, driving $E$ to mark the regions the decision relies on. In Stage 2, the frozen detector conditions the Evidence-Guided Explainer (\textbf{EGE}) on $E$ to generate the explanation under contrastive decoding.}
\label{fig:framework}
\end{figure*}

\section{Related Work}
\subsection{Generalizable AI-Generated Image Detection}
Existing research on AIGI detection has largely centered on cross-generator generalization along three lines. The first extracts low-level forensic traces tied to the generation process, such as diffusion reconstruction error~\cite{wang2023dire,ricker2024aeroblade}, frequency and up-sampling artifacts~\cite{tan2024frequency,tan2024npr}, and spectral cues~\cite{karageorgiou2025any}. NPR~\cite{tan2024npr}, for instance, shows that up-sampling leaves traces general enough to transfer from ProGAN alone. The second adapts large pre-trained vision models, ranging from linear probes on frozen CLIP features~\cite{ojha2023univfd} to prompt and adapter-based variants that steer the representation toward forgery cues~\cite{tan2025c2pclip,chen2025forgelens}. The third fuses low-level artifacts with high-level semantics for complementary generalization~\cite{yan2025aide,cheng2025cospy}. Across these lines, optimization targets overall detection performance. Shortcut analyses further show that high accuracy on some benchmarks may reflect content or format bias rather than genuine generation traces~\cite{guillaro2025bias}, and whether the detector's decision depends on the artifacts it extracts remains untested.

\subsection{Explainable Analysis of AI-Generated Image}
Multimodal large language models have extended detection from a bare verdict to a human-readable explanation along two lines. The first predicts the verdict and its explanation jointly within a single framework~\cite{huang2025sida,xu2025fakeshield,kang2025legion,wen2025fakevlm,ji2026fakexplain}. The second decouples them, feeding detection-stage cues such as attention maps, forgery scores, or textual clues into an LLM to support its reasoning~\cite{guo2025m2f2,chen2025x2dfd,yu2025unlocking,zhou2025aigi}. Neither line, however, explicitly tests whether the cues an explanation cites are the ones the prediction actually rests on. Attention maps, for instance, reveal where a model reads rather than what drives its decision~\cite{jain2019attention}, so an explanation may read as plausible yet remain unfaithful to the underlying evidence.

Aware of this gap, recent work has begun to evaluate the hallucination and faithfulness of forensic explanations. DeepfakeJudge~\cite{kuckreja2026deepfakejudge} treats reasoning fidelity as a quantifiable dimension via an MLLM-as-a-Judge, yet remains within an \emph{MLLM-evaluating-MLLM} paradigm and primarily measures human-aligned plausibility. TriDF~\cite{jiang2026tridf} jointly benchmarks perception, detection, and hallucination, revealing that hallucination remains severe even for state-of-the-art MLLMs. EviDet complements these approaches by grounding explanations in evidence the detector relies on, and we adopt TriDF's hallucination metrics in our experiments.

\section{Methodology}
We present EviDet, a two-stage framework built around a shared patch-level evidence map $E$ (Figure~\ref{fig:framework}). \textbf{Stage 1} learns a decision-relevant evidence map $E$ inside the detector: \textbf{MSRP} provides low-level artifacts and DINOv3 supplies self-supervised features for constructing it, while \textbf{EMV} shapes $E$ through keep and drop interventions. \textbf{Stage 2} freezes the detector and grounds the explanation in $E$ through \textbf{EGE}, so one map serves both detection and explanation.

\subsection{Problem Formulation}
Given an image $x$, EviDet outputs a three-way prediction $\hat{y} \in \{\text{real}, \text{fully synthetic}, \text{tampered}\}$, a patch-level evidence map \(E \in [0,1]^{32\times32}\) whose entry \(e_i \in [0,1]\) scores how strongly patch \(i\) supports a fake decision, and a natural-language explanation \(T\). A score \(e_i\) counts as \emph{evidence} only when it (i) enters the classifier's forward pass and (ii) shifts the fake response under feature-level intervention. Neither condition alone suffices, and our analysis verifies both. This sets $E$ apart from artifact responses, attention, post-hoc activation maps, and manipulation masks~\cite{jain2019attention,selvaraju2017grad}, none of which is both used by the detector and verified to be so. Grounding the explanation in such decision-relevant evidence is what makes it faithful.

\subsection{Multi-Scale Resampling Prior (MSRP)}
Up-sampling operations, common to GAN and diffusion pipelines, leave local traces that transfer across generators, and NPR~\cite{tan2024npr} exploits them through a fixed $2\times2$ differencing pattern. This hard-coded descriptor, however, commits to one spatial form and scale of the artifact, so it degrades once generators change the up-sampling operator or strengthen post-upsampling refinement.

\textbf{MSRP} therefore encodes the resampling process rather than one fixed signature. For a set of scales \(\mathcal{S}=\{0.60,0.75,0.90\}\), we downscale and restore the input image back to the original resolution to obtain \(x^{(s)}\), and apply the SRM high-pass operator \(H(\cdot)\)~\cite{fridrich2012rich} to both \(x\) and \(x^{(s)}\). The response map averages their high-frequency residuals across scales:
\begin{equation}
    R = \frac{1}{|\mathcal{S}|}\sum_{s\in\mathcal{S}} \bigl|H(x)-H(x^{(s)})\bigr|.
\label{eq:msrp}
\end{equation}
Averaging reduces sensitivity to a single resampling operator, while consistent high-frequency shifts under resampling remain visible in \(R\).

The response map $R$ obtained above is a candidate artifact, not a detection score. Across the three classes its statistics differ in a stable, class-dependent way (full statistics in Appendix~A), so the discriminative signal lies in the spatial organization of the response rather than its magnitude. A raw threshold on $R$ would therefore be unreliable. The artifact must instead be selected through fusion with richer visual features.

\subsection{From Artifact Cues to an Evidence Map}
We combine the low-level artifact cues in $R$ with high-level visual features to learn a spatial evidence map $E$. The input image $x$, preprocessed to $512 \times 512$, is encoded by a DINOv3 ViT-L/16 adapted with LoRA~\cite{hu2022lora}, yielding patch features $F_{\mathrm{enc}} \in \mathbb{R}^{N \times C}$ on a $32 \times 32$ grid. We align $R$ to the same grid and project it into token space, obtaining $F_{\mathrm{res}} \in \mathbb{R}^{N \times C}$. The fusion module then combines the two token sets through cross-attention, producing one output token for each position on the query grid:
\begin{equation}
    \widetilde{F} = \mathcal{F}_{\mathrm{fuse}}\!\left( Q = F_{\mathrm{enc}},\; K = V = F_{\mathrm{res}} \right).
    \label{eq:evidence_fusion}
\end{equation}
Each fused patch feature $\widetilde{f}_i$ in $\widetilde{F}$ is scored by a shared MLP head:
\begin{equation}
e_i = \sigma\bigl(\mathrm{MLP}_{e}(\widetilde{f}_i)\bigr) \in [0,1],
\end{equation}
and the per-patch scores $\{e_i\}_{i=1}^{N}$ form the spatial evidence map $E$, which weights the encoder features
pooled for classification:
\begin{equation}
F_{det} = \frac{\sum_i (1+e_i)\,f^{\rm enc}_i}{\sum_i (1+e_i)},
\qquad
\hat{y} = \arg\max \mathrm{MLP}_{\rm cls}(F_{det}),
\label{eq:gated-pooling}
\end{equation}
where $f^{\rm enc}_i$ is the $i$-th patch feature of $F_{\rm enc}$. $R$ influences the decision only through $E$, which reweights the encoder features while retaining all patches. The three-class cross-entropy $\mathcal{L}_{\mathrm{det}}$ trains the classifier and $E$ jointly, with gradients reaching each $e_i$ through the normalized pooling in Eq.~\eqref{eq:gated-pooling}.

However, $\mathcal{L}_{\text{det}}$ provides $E$ no explicit spatial target. Manipulation masks supply such targets only for tampered images, while real and fully synthetic images have class labels alone. To limit shortcut learning under this asymmetry, we add a soft auxiliary class-conditioned regularizer $\mathcal{L}_{\mathrm{evi}}$. On real images, $E$ is suppressed because it represents forgery cues, not authenticity. On tampered images, the mask serves as soft support rather than a dense segmentation target, since discriminative cues concentrate along manipulation boundaries. On fully synthetic images, a ranking margin favors high-MSRP patches over low-MSRP ones, using $R$ as a coarse spatial prior (exact forms and weights in Appendix~B). These targets balance the learning signal across classes.

\subsection{Evidence Masking Verification (EMV)}
Evidence-weighted pooling and $\mathcal{L}_{\mathrm{evi}}$ shape $E$ per class, yet neither ensures the detector relies on the highlighted patches.  \textbf{EMV} supplies this missing constraint. For fake images that the original branch predicts as fake, we impose two criteria on $E$: retaining the highlighted patches should sustain the fake response, while suppressing them should reduce it. To avoid pixel masking or cropping artifacts, \textbf{EMV} applies these interventions directly on the encoder features $F_{\mathrm{enc}}$ used for classification (Figure~\ref{fig:framework}).

During training, we form a soft visibility mask from the current $E$: $M_i=\sigma\bigl((e_i-\bar{e})/\tau\bigr)$, where $\bar{e}$ is the mean evidence score and the temperature $\tau$ is annealed downward to sharpen the mask. We detach $M$ to fix the mask within each update, while keeping the pooling weights $(1+e_i)$ differentiable. The \textbf{keep branch} uses $M\odot F_{\mathrm{enc}}$, and the \textbf{drop branch} uses $(1-M)\odot F_{\mathrm{enc}}$. Both apply the evidence-weighted pooling of Eq.~\eqref{eq:gated-pooling} with its original denominator and share the classification head. We optimize their responses jointly:
\begin{equation}
\mathcal{L}_{\rm emv}=
\underbrace{-\log P^{\rm keep}_{\rm fake}}_{\mathcal{L}_{\rm keep}}
+\underbrace{\max\bigl(0,\,m-(P^{\rm ori}_{\rm fake}-P^{\rm drop}_{\rm fake})\bigr)}_{\mathcal{L}_{\rm drop}},
\label{eq:emv}
\end{equation}
where $P_{\mathrm{fake}}$ sums the fully synthetic and tampered probabilities and $m$ sets the target drop in fake confidence. The two terms promote sufficiency and comprehensiveness for the fake response. Centered at $\bar{e}$, a uniform $E$ gives $M_i=0.5$ without spatial selection, leaving $\mathcal{L}_{\mathrm{emv}}$ with a positive lower bound under the given loss and margin $m$ (Appendix~C). \textbf{EMV} is training-only, adding no inference overhead.

The full Stage~1 objective is $\mathcal{L}_{\mathrm{stage1}} = \mathcal{L}_{\mathrm{det}} + \lambda_e\mathcal{L}_{\mathrm{evi}} + \lambda_v\mathcal{L}_{\mathrm{emv}}$, where $\mathcal{L}_{\mathrm{det}}$ supervises three-way classification, $\mathcal{L}_{\mathrm{evi}}$ shapes $E$, and $\mathcal{L}_{\mathrm{emv}}$ grounds the fake decision in it.

\subsection{Evidence-Guided Explainer (EGE)}
\label{sec:ege}

Stage~2 uses the evidence map $E$ learned with \textbf{EMV} to guide explanation generation. As decoding proceeds, the influence of the visual condition gradually weakens and the language prior becomes dominant, a shift shown to correlate strongly with hallucination~\cite{favero2024m3id}. In forensic explanation, this can lead to generic claims, such as ``inconsistent lighting,'' without sufficient image-specific support. \textbf{EGE} mitigates this by conditioning on the detector's features and explicitly encouraging generation to depend on $E$.

\textbf{EGE} conditions generation only on what the frozen detector provides. We reuse its patch features $F_{\rm enc}$ and the evidence map $E$, project them into the token space of a large language model without passing through its vision tower, and form a continuous prefix together with the predicted class and a task prompt. However, supervised fine-tuning alone lets the model fit reference explanations while ignoring $E$, \textbf{EGE} therefore contrasts two paths that share $F_{\rm enc}$, the predicted class, and the prompt but differ only in $E$. The \emph{true} path uses the $E$ and the \emph{null} path zeroes it, so any gain of the true path over the null path is attributable to the $E$.

\begin{table*}[t]
\centering
\small
\setlength{\tabcolsep}{10.5pt}
\begin{tabular}{cccccccccc}
\toprule
\multirow{2}{*}{Method} & \multirow{2}{*}{Venue} & \multicolumn{2}{c}{Real} & \multicolumn{2}{c}{Fully Synthetic} & \multicolumn{2}{c}{Tampered} & \multicolumn{2}{c}{Overall} \\
\cmidrule(lr){3-4} \cmidrule(lr){5-6} \cmidrule(lr){7-8} \cmidrule(lr){9-10}
& & Acc & F1 & Acc & F1 & Acc & F1 & Acc & F1 \\
\midrule
Gram-Net & CVPR'20 & 89.2 & 91.7 & 97.9 & 98.6 & 89.9 & 86.9 & 95.3 & 92.4 \\
LGrad & CVPR'23 & 62.0 & 76.1 & 58.0 & 66.3 & \underline{99.1} & \underline{98.8} & 73.0 & 80.4 \\
UnivFD & CVPR'23 & 68.3 & 68.5 & 86.4 & 98.0 & 92.5 & 90.0 & 82.4 & 85.5 \\
NPR & CVPR'24 & 86.0 & 91.5 & 97.6 & 96.8 & 95.2 & 93.0 & 92.9 & 93.8 \\
LLaVA-OV-7B$\dagger$ & Arxiv'24 & \textbf{99.6} & 50.6 & 2.0 & 3.8 & 1.3 & 2.6 & 34.4 & 19.0 \\
Qwen3-VL-8B$\dagger$ & Arxiv'25 & 97.4 & 55.2 & 37.0 & 53.8 & 4.0 & 7.4 & 46.2 & 38.8 \\
D$^3$ & CVPR'25 & 85.7 & 88.8 & 96.9 & 96.1 & 91.7 & 89.3 & 91.4 & 91.4 \\
SIDA-7B & CVPR'25 & 89.1 & 91.0 & \underline{98.7} & 98.6 & 91.2 & 91.0 & 93.5 & 93.5 \\
SIDA-13B & CVPR'25 & 89.6 & 91.1 & 98.5 & \underline{98.7} & 92.9 & 91.2 & 93.6 & 93.5 \\
MaskCLIP & CVPR'25 & 95.4 & \underline{96.2} & 98.1 & 98.3 & 97.5 & 96.7 & \underline{97.1} & \underline{97.3} \\
\midrule
\textbf{EviDet (ours)} & \multicolumn{1}{c}{--} & \underline{98.8} & \textbf{99.1} & \textbf{100.0} & \textbf{99.9} & \textbf{99.5} & \textbf{99.2} & \textbf{99.4} & \textbf{99.4} \\
\bottomrule
\end{tabular}
\caption{Comparison with state-of-the-art methods on the official SID-Set split. We report Accuracy and  F1 (\%) across three fine-grained classes, $\dagger$ denotes zero-shot evaluation. Best and second-best results are in \textbf{bold} and \underline{underlined}, respectively.}
\label{tab:sidset}
\end{table*}

\begin{table*}[t]
\centering
\small
\setlength{\tabcolsep}{5.4pt}
\begin{tabular}{cccccccccccccc}
\toprule
\multirow{2}{*}{Method} & \multirow{2}{*}{Venue} & \multicolumn{2}{c}{SD1.5} & \multicolumn{2}{c}{SD2.1} & \multicolumn{2}{c}{SDXL} & \multicolumn{2}{c}{SD3} & \multicolumn{2}{c}{Flux.1} & \multicolumn{2}{c}{AVG} \\
\cmidrule(lr){3-4} \cmidrule(lr){5-6} \cmidrule(lr){7-8} \cmidrule(lr){9-10} \cmidrule(lr){11-12} \cmidrule(lr){13-14}
& & F1 & Acc & F1 & Acc & F1 & Acc & F1 & Acc & F1 & Acc & F1 & Acc \\
\midrule
ObjectFormer & CVPR'22 & 71.7 & 75.2 & 66.8 & 72.6 & 49.2 & 62.9 & 48.3 & 62.5 & 37.9 & 58.1 & 54.8 & 66.3 \\
UnivFD & CVPR'23 & 77.5 & 77.6 & 80.6 & 81.9 & 70.7 & 74.8 & 71.1 & 75.2 & 61.1 & 69.1 & 72.2 & 75.7 \\
IML-ViT & AAAI'24 & \underline{94.5} & 75.7 & 69.7 & 61.2 & 41.0 & 50.0 & 44.7 & 51.3 & 18.2 & 43.6 & 53.6 & 56.4 \\
FreqNet & AAAI'24 & 75.9 & 77.7 & 61.0 & 68.4 & 53.2 & 64.0 & 53.5 & 64.4 & 38.5 & 57.1 & 56.4 & 66.3 \\
NPR & CVPR'24 & 79.4 & 79.3 & 81.7 & 81.8 & 72.1 & 74.3 & 73.4 & 75.5 & 67.6 & 71.4 & 74.9 & 76.5 \\
RINE & ECCV'24 & 91.1 & 91.0 & 87.5 & 88.1 & 73.4 & 78.8 & 72.1 & 76.8 & 55.9 & 67.0 & 76.0 & 80.3 \\
MaskCLIP & CVPR'25 & 92.6 & \underline{92.7} & \underline{88.7} & \underline{89.5} & 78.0 & 81.2 & 73.1 & 78.0 & 56.5 & 68.5 & 77.8 & 82.0 \\
FTNet-T & AAAI'26 & 78.0 & 77.7 & 84.1 & 84.8 & \underline{85.8} & \underline{86.2} & \underline{84.1} & \underline{84.7} & \underline{81.4} & \underline{82.5} & \underline{82.7} & \underline{83.2} \\
\midrule
\textbf{EviDet (ours)} & \multicolumn{1}{c}{--} & \textbf{99.8} & \textbf{99.7} & \textbf{99.7} & \textbf{99.6} & \textbf{99.5} & \textbf{99.3} & \textbf{99.5} & \textbf{99.3} & \textbf{94.0} & \textbf{92.4} & \textbf{98.5} & \textbf{98.1} \\
\bottomrule
\end{tabular}
\caption{Generalization on OpenSDID. F1 and Accuracy (\%) of detectors trained on SD1.5 and evaluated on unseen diffusion generators over three random seeds. Best and second-best are marked in \textbf{bold} and \underline{underline}, respectively.}
\label{tab:opensdi}
\end{table*}

At inference, for images predicted as fake, both paths are evaluated at every decoding step, yielding next-token distributions $p_{\rm true}$ and $p_{\rm null}$. The score
\begin{equation}
s(v)=\log p_{\rm true}(v)+\alpha\bigl(\log p_{\rm true}(v)-\log p_{\rm null}(v)\bigr)
\label{eq:cd}
\end{equation}
is then used to re-score the top candidates of the true path, where $v$ denotes a candidate next token and $\alpha>0$ controls the strength of the evidence contrast. The first term retains the true-path preference, while the contrastive term favors tokens whose probability increases when $E$ is provided. The same contrast shapes training. Beyond a standard supervised fine-tuning loss $\mathcal{L}_{\rm sft}$ for forensic vocabulary, we apply the margin loss $\mathcal{L}_{\rm ec}=\max\bigl(0,\,m-(S_{\rm true}-S_{\rm null})\bigr)$ to fake samples with weight $\lambda_{\rm ec}$ to encourage a higher likelihood of the reference explanation under the true path than under the null path. Here, $S_{\rm true}$ and $S_{\rm null}$ denote the log-likelihoods of the same reference explanation under the two paths, respectively.

Overall, detection, verification, and explanation are unified around a single evidence map $E$: the detector decides through it, \textbf{EMV} encourages reliance on it, and \textbf{EGE} grounds the explanation on it.

\section{Experiments}

\subsection{Experimental Setup}
\noindent\textbf{Datasets. }Our primary benchmark is SID-Set~\cite{huang2025sida}, with 100K real, 100K fully synthetic, and 100K tampered social-media images, tampered-region masks, and about 3K reference explanations. We follow the official split and evaluation protocol. To assess cross-generator generalization, we further evaluate on OpenSDID, the dataset introduced by the OpenSDI benchmark~\cite{wang2025opensdi}, training on SD1.5 and testing on the unseen generators SD2.1, SDXL, SD3, and Flux.1. Explanation quality is evaluated on MMTD-Set~\cite{xu2025fakeshield} and on the official evaluation set of TriDF~\cite{jiang2026tridf} (image track), following its prescribed hallucination assessment protocol. For methods evaluated under the same protocol, we report the results from their original papers to ensure a fair comparison.

\noindent\textbf{Baselines.} We compare EviDet against a comprehensive set of state-of-the-art methods. For fine-grained detection on SID-Set, we compare our method with Gram-Net~\cite{liu2020global}, LGrad~\cite{tan2023learning}, NPR~\cite{tan2024npr}, UnivFD~\cite{ojha2023univfd}, D$^3$~\cite{yang2025d}, MaskCLIP~\cite{wang2025opensdi}, zero-shot MLLMs LLaVA-OneVision-7B~\cite{li2024llava}, Qwen3-VL-8B~\cite{bai2025qwen3}, and use SIDA-7B/13B~\cite{huang2025sida} as the native fine-grained baselines. For cross-generator generalization on OpenSDID, we further compare with ObjectFormer~\cite{wang2022objectformer}, IML-ViT~\cite{ma2023imlvit}, FreqNet~\cite{tan2024frequency}, RINE~\cite{koutlis2024leveraging}, and FTNet~\cite{yao2026leveraging} under the official protocol. For explanation quality, we compare with InternVL3.5-8B~\cite{wang2025internvl3}, Qwen3-VL-8B~\cite{bai2025qwen3}, LLaVA-OneVision-7B~\cite{li2024llava}, MiMo-VL-7B~\cite{li2025xiaomi}, FakeShield~\cite{xu2025fakeshield}, FakeVLM~\cite{wen2025fakevlm}, an image-input LoRA-tuned Qwen3-VL, and the description branch of SIDA.

\noindent\textbf{Evaluation Metrics. }We evaluate detection with image-level accuracy and F1. Explanations are evaluated by ROUGE-L and cosine semantic similarity (CSS) against reference texts, following~\cite{xu2025fakeshield, kang2025legion}, and by the TriDF hallucination metrics (Cover, CHAIR, Hal, and F$_{0.5}$)~\cite{jiang2026tridf}, which score explanations against human-annotated artifacts under the official protocol. The decision relevance of $E$ is measured by sufficiency and comprehensiveness~\cite{deyoung2020eraser}, adapted from NLP rationale evaluation to image patches.

\noindent\textbf{Implementation Details. }We implement EviDet in two stages: Stage 1 learns the evidence map and the detector, and Stage 2 freezes the detector and trains the evidence-guided explainer. In Stage 1, images are square-cropped on the short side, resized to $512\times512$, and encoded by a frozen DINOv3 ViT-L/16 with LoRA adapters for $10$ epochs. \textbf{MSRP} response maps are precomputed at the same resolution over scales $\{0.60, 0.75, 0.90\}$. In Stage 2, the explainer trains a projector and rank-16 LoRA adapters on Qwen3-VL-8B for $5$ epochs, using evidence-null contrastive decoding with $\alpha=1.0$ and top-$k=50$. All experiments are conducted on two NVIDIA RTX 3090 GPUs. (Full optimization and hyper-parameter settings are provided in Appendix~D).

\subsection{Detection Performance}
\subsubsection{Comparison with State of the Art}
Table~\ref{tab:sidset} reports three-class detection results on SID-Set. EviDet achieves 99.4\% overall accuracy and F1, exceeding MaskCLIP by 2.3/2.1 points and SIDA-13B by 5.8/5.9 points, and maintaining F1 above 99\% in every class rather than peaking on fully synthetic images alone. Zero-shot MLLMs instead bias toward real predictions, as LLaVA-OV scores 99.6\% on real ones but 2.0\% and 1.3\% on the fake classes. This balanced gain across classes aligns with our joint design: \textbf{MSRP} complements visual features with low-level cues, \textbf{EMV} trains selected features to support fake predictions, and class-conditioned regularization extends spatial guidance to all three classes.

\subsubsection{Robustness}
Trained on clean data and tested under four perturbations (JPEG, blur, noise, resize), the purely low-level NPR collapses once pixel statistics are disturbed (Macro-F1 from 93.8\% to 41.2\% under JPEG q50), whereas EviDet retains 93.8\% under the same compression and attains the highest average F1 across all levels. (Full results in Appendix~E).

\subsubsection{Generalization Capability Evaluation}
Table~\ref{tab:opensdi} evaluates cross-generator generalization on OpenSDID. EviDet surpasses the strongest average baseline, the 8-shot FTNet-T, by 14.9\% Acc and 15.8\% F1, and maintains \textbf{97.7\%} Acc and \textbf{98.2\%} F1 on the four unseen generators, where FTNet-T reaches 84.5\% Acc. This stability across SD2.1--SD3 stems from \textbf{MSRP} capturing generic rescaling sensitivity rather than generator-specific fingerprints, paired with \textbf{EMV}-guided localization. Even on Flux.1, the hardest unseen generator, EviDet retains 94.0\% F1 against MaskCLIP's 56.5\%, with error inspection showing attenuated \textbf{MSRP} responses only on ultra-smooth images lacking fine texture.

\subsection{Decision Relevance of the Evidence Map}
To test whether predictions depend on $E$, we follow~\cite{deyoung2020eraser} and report the mean $P_{\mathrm{fake}}$ drop when keeping only top-ranked patches (sufficiency) or dropping them (comprehensiveness). All maps are evaluated on the same frozen \textbf{EMV}-trained detector under a shared patch budget matched to $E$'s selected area ($E$ w/o \textbf{EMV} comes from a detector trained without \textbf{EMV}). For Random and $E$ w/o \textbf{EMV}, keeping selected patches preserves fake confidence, while dropping them changes little. By contrast, $E$ pairs low Suff.\ (0.065) with high Comp.\ (0.788) for the largest $\Delta$ (0.723, Table~\ref{tab:evidence_audit}). This sharper keep/drop contrast confirms the decision relevance of \textbf{EMV}-trained maps over attention and Grad-CAM (see Appendix~F for fixed-budget checks).

\begin{table}[htbp]
\centering
\small
\setlength{\tabcolsep}{6.5pt}
\begin{tabular}{cccc}
\toprule
Ranking map & Suff. $\downarrow$ & Comp. $\uparrow$ & $\Delta$ $\uparrow$\\
\midrule
Random (3 seeds) & 0.062 & 0.054 & -0.008 \\
Inverted $E$ & 0.791 & 0.069 & -0.722 \\
MSRP response $R$ & 0.117 & 0.508 & 0.391 \\
DINOv3 attention map & 0.074 & 0.505 & 0.431 \\
Grad-CAM & 0.263 & 0.176 & -0.087 \\
\midrule
$E$ w/o EMV & 0.007 & 0.040 & 0.033 \\
\textbf{$E$ (ours)} & 0.065 & 0.788 & \textbf{0.723} \\
\bottomrule
\end{tabular}
\caption{Intervention comparison of ranking maps on SID-Set. A decision-relevant map must yield both low Sufficiency and high Comprehensiveness ($\Delta$ = Comp.$-$Suff.)}
\label{tab:evidence_audit}
\end{table}

Localization further supports $E$'s spatial alignment: upsampling it to mask resolution yields 99.1\% AUC, 75.0\% F1, and 64.2\% IoU on SID-Set, surpassing dedicated methods such as SIDA-7B and MaskCLIP (Appendix~F).

\subsection{Explanation Performance}
\noindent\textbf{Reference Alignment.} As shown in Table~\ref{tab:explanation-reference}, EviDet achieves 42.11 ROUGE-L and 82.23 CSS, outperforming SIDA by 1.1 and 3.5 points on SID-Set. This larger CSS gain shows that conditioning on the evidence map strengthens semantic content rather than surface phrasing alone. On zero-shot transfer to MMTD-Set, EviDet also leads on CSS (70.05, +2.4 over SIDA) while remaining competitive on ROUGE-L.

\begin{table}[htbp]
\centering
\small
\setlength{\tabcolsep}{6pt}
\begin{tabular}{ccccc}
\toprule
\multirow{2}{*}{Method} & \multicolumn{2}{c}{SID-Set} & \multicolumn{2}{c}{MMTD-Set} \\
\cmidrule(lr){2-3} \cmidrule(lr){4-5}
& R-L$\uparrow$ & CSS$\uparrow$ & R-L$\uparrow$ & CSS$\uparrow$ \\
\midrule
InternVL3.5-8B & 20.22 & 64.77 & 18.94 & 58.63 \\
Qwen3-VL-8B & 18.69 & 66.02 & 16.65 & 57.88 \\
LLaVA-OV-7B & 17.77 & 61.06 & 17.39 & 58.59 \\
MiMo-VL-7B & 19.66 & 63.36 & 19.58 & 61.57 \\
\midrule
Qwen3-VL-8B (SFT) & 38.86 & 77.69 & \underline{23.03} & 61.49 \\
SIDA-7B & \underline{41.03} & \underline{78.74} & \textbf{23.55} & \underline{67.63} \\
\textbf{EviDet (ours)} & \textbf{42.11} & \textbf{82.23} & 22.18 & \textbf{70.05} \\
\bottomrule
\end{tabular}
\caption{Reference-based explanation result. Upper block: zero-shot MLLMs. Lower block: models fine-tuned on SID-Set only, with MMTD-Set evaluated without further tuning. ROUGE-L (R-L) and CSS are multiplied by 100.}
\label{tab:explanation-reference}
\end{table}

\noindent\textbf{Hallucination.} Table~\ref{tab:tridf} evaluates explanation reliability against human-annotated artifacts on TriDF. Removing \textbf{EMV} from evidence learning lowers Acc to 0.650 and raises Hal to 0.964 even with CD enabled. With the \textbf{EMV}-trained $E$ fixed, enabling CD further reduces CHAIR from 0.726 to 0.540 while raising Cover from 0.247 to 0.385. These ablations support the contributions of evidence learning and contrastive decoding to explanation reliability.

\begin{table}[t]
\centering
\small
\setlength{\tabcolsep}{4pt}
\begin{tabular}{cccccc}
\toprule
Method & Acc$\uparrow$ & Cover$\uparrow$ & CHAIR$\downarrow$ & Hal$\downarrow$ & F$_{0.5}$$\uparrow$ \\
\midrule
InternVL3.5-8B & 0.418 & 0.027 & 0.975 & 1.000 & 0.030 \\
Qwen3-VL-8B & 0.621 & \underline{0.256} & 0.807 & 0.999 & 0.202 \\
LLaVA-OV-7B & 0.385 & 0.000 & 1.000 & 1.000 & 0.003 \\
MiMo-VL-7B & 0.565 & 0.228 & \underline{0.654} & \underline{0.874} & \underline{0.291} \\
\midrule
FakeShield & 0.405 & 0.025 & 0.975 & 0.997 & 0.031 \\
FakeVLM & 0.474 & 0.006 & 0.995 & 1.000 & 0.005 \\
SIDA-7B & \underline{0.663} & 0.090 & 0.856 & 0.990 & 0.122 \\
EviDet w/o EMV  & 0.650 & 0.210 & 0.743 & 0.964 & 0.241 \\
EviDet w/o CD & \textbf{0.705} & 0.247 & 0.726 & 0.933 & 0.261 \\
\textbf{EviDet (ours)} & \textbf{0.705} & \textbf{0.385} & \textbf{0.540} & \textbf{0.840} & \textbf{0.424} \\
\bottomrule
\end{tabular}
\caption{Zero-shot explanation reliability on TriDF (Upper: open-source MLLMs, Lower: forgery methods).}
\label{tab:tridf}
\end{table}

\subsection{Ablation Study}
Table~\ref{tab:ablation} ablates EviDet's detection components on SID-Set under a fixed split and training budget. Single branches fall short on their own: a frozen DINOv3 linear probe reaches 82.3\%, whereas pairing it with \textbf{MSRP} after LoRA tuning reaches 92.6\%. Adding EMV-keep or EMV-drop alone yields moderate gains (95.5\% and 95.7\%), while combining both jumps to 99.4\%. Specifically, joint \textbf{EMV} optimization enforces complementary constraints on $E$: keep retains decision-relevant cues and drop penalizes over-broad maps, improving sufficiency and comprehensiveness together (Additional ablations are provided in Appendix~G and ~H).

\begin{table}[htbp]
\centering
\small
\setlength{\tabcolsep}{5.5pt}
\begin{tabular}{cccccc}
\toprule
\multicolumn{4}{c}{Module} & \multirow{2}{*}{Acc} & \multirow{2}{*}{F1} \\
\cmidrule(lr){1-4}
DINO & MSRP & EMV-keep & EMV-drop & & \\
\midrule
\checkmark(frozen) & & & & 82.3 & 82.7 \\
\checkmark & & & & 88.2 & 89.4 \\
& \checkmark & & & 83.8 & 84.9 \\
\checkmark & \checkmark & & & 92.6 & 92.5 \\
\checkmark & \checkmark & \checkmark & & 95.5 & 94.9 \\
\checkmark & \checkmark & & \checkmark & 95.7 & 95.3 \\
\midrule
\checkmark & \checkmark & \checkmark & \checkmark & \textbf{99.4} & \textbf{99.4} \\
\bottomrule
\end{tabular}
\caption{Ablation of the detection stage on SID-Set. Accuracy and F1 (\%) over three classes. Best results are in \textbf{bold}.}
\label{tab:ablation}
\end{table}

\subsection{Qualitative Results}
\begin{figure}[htbp]
\centering
\includegraphics[width=0.9\columnwidth]{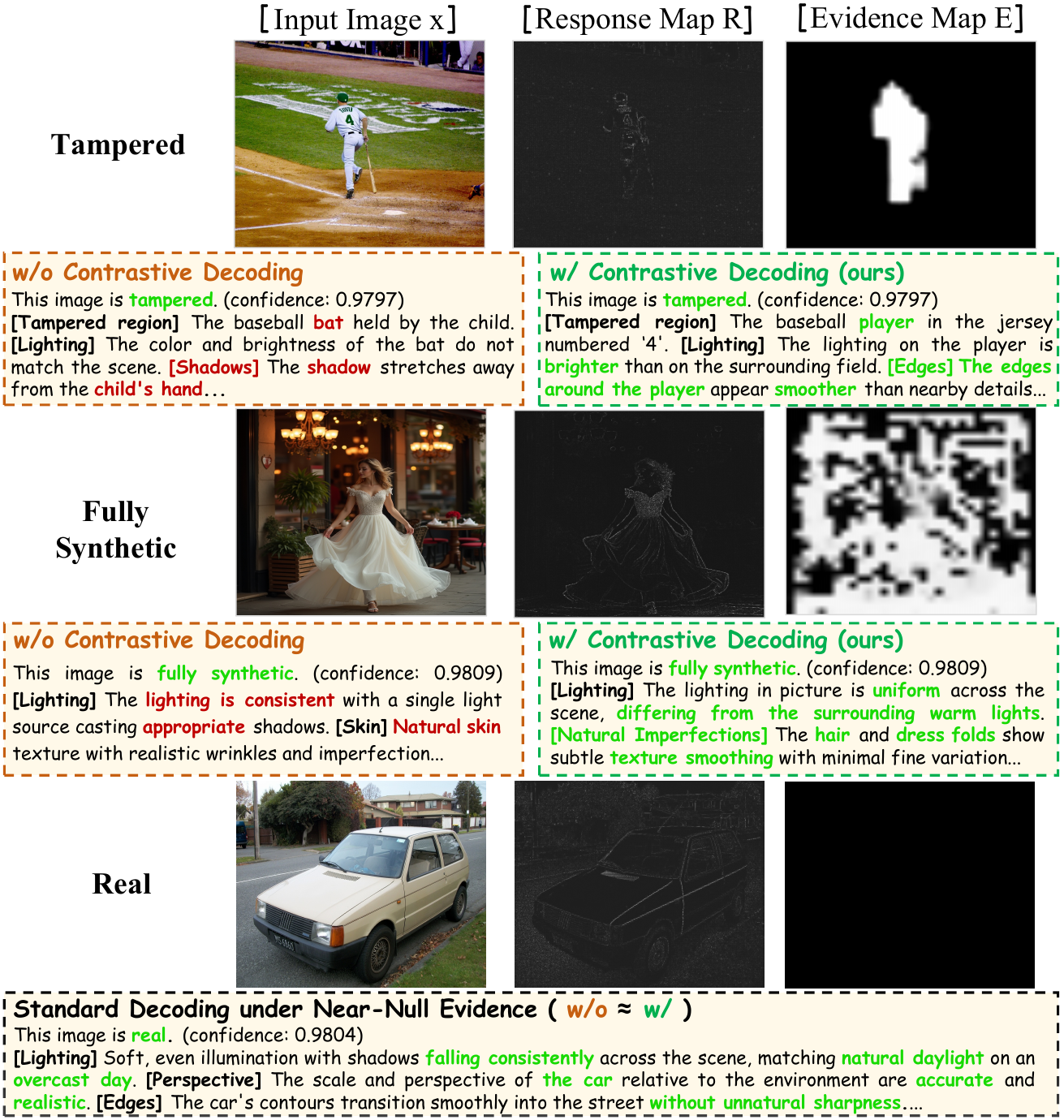} 
\caption{Qualitative results on SID-Set. $E$ behaves distinctly across the three classes and the highlighted text marks where contrastive decoding corrects generic claims into evidence-specific findings.}
\label{fig:qualitative}
\end{figure}

As shown in Figure~\ref{fig:qualitative}, using one representative example per class, we visualize the response map $R$, the evidence map $E$ (upsampled for visualization), and the explanations from standard decoding (fine-tuning only) and our evidence-guided contrastive decoding (\textbf{EGE}). The two schemes diverge on the same fake image: standard decoding falls back on priors from fine-tuning, producing phrases such as ``natural skin'' that contradict its own verdict, as the visual condition carried by $E$ is diluted by the language prior over a long generation. \textbf{EGE} keeps generation anchored to E, yielding evidence-specific findings such as the edited player in the tampered example, and directing attention to background and structural cues rather than the semantically dominant object alone. This is the faithful behavior our design targets, consistent with the hallucination reduction in Table~\ref{tab:tridf} (full results, more examples, and failure cases in Appendix~I).

\section{Conclusion}
Explanations produced by existing methods can be plausible but unfaithful, as it remains unclear whether they reflect the evidence underlying the decision. We presented EviDet, which addresses this gap by connecting detection and explanation through a single patch-level evidence map $E$. In the detection stage, \textbf{MSRP} extracts multi-scale resampling residuals, which are combined with visual features to learn $E$, and \textbf{EMV} encourages its decision relevance through keep/drop interventions. In the explanation stage, \textbf{EGE} conditions a language model on this map through evidence-null contrastive training and decoding, which lowers the rate of hallucinated artifacts. Experiments on SID-Set, OpenSDID, and TriDF show gains in fine-grained accuracy, cross-generator generalization, and explanation reliability. These findings support the use of the shared evidence map $E$ to connect detection and explanation, offering a practical approach to generating forensic explanations that more faithfully reflect the evidence behind the detector's decisions.

In future work, we will curate diverse, finely annotated datasets with texture-grounded explanations for fine-tuning. On the inference side, we will distill the evidence contrast into a single-path model to reduce the extra decoding cost.

\bibliography{EviDet}


\setcounter{secnumdepth}{2} 

%
\clearpage
\twocolumn[
  \begin{center}
    \LARGE \textbf{Supplementary Material}
  \end{center}
  \vspace{3em}
]
\appendix
\numberwithin{figure}{section}
\numberwithin{table}{section}
\numberwithin{equation}{section}

\section{Multi-Scale Resampling Prior (MSRP) Details and Candidate-Cue Statistics}
\label{app:msrp}

This section visualizes how \textbf{MSRP} constructs the response map $R$ and examines its behavior across real, fully synthetic, and tampered
images. We first illustrate the multi-scale resampling process, then compare representative response patterns and their class-conditional statistics on SID-Set.

\begin{figure*}[htbp]
\centering
\includegraphics[width=0.84\textwidth]{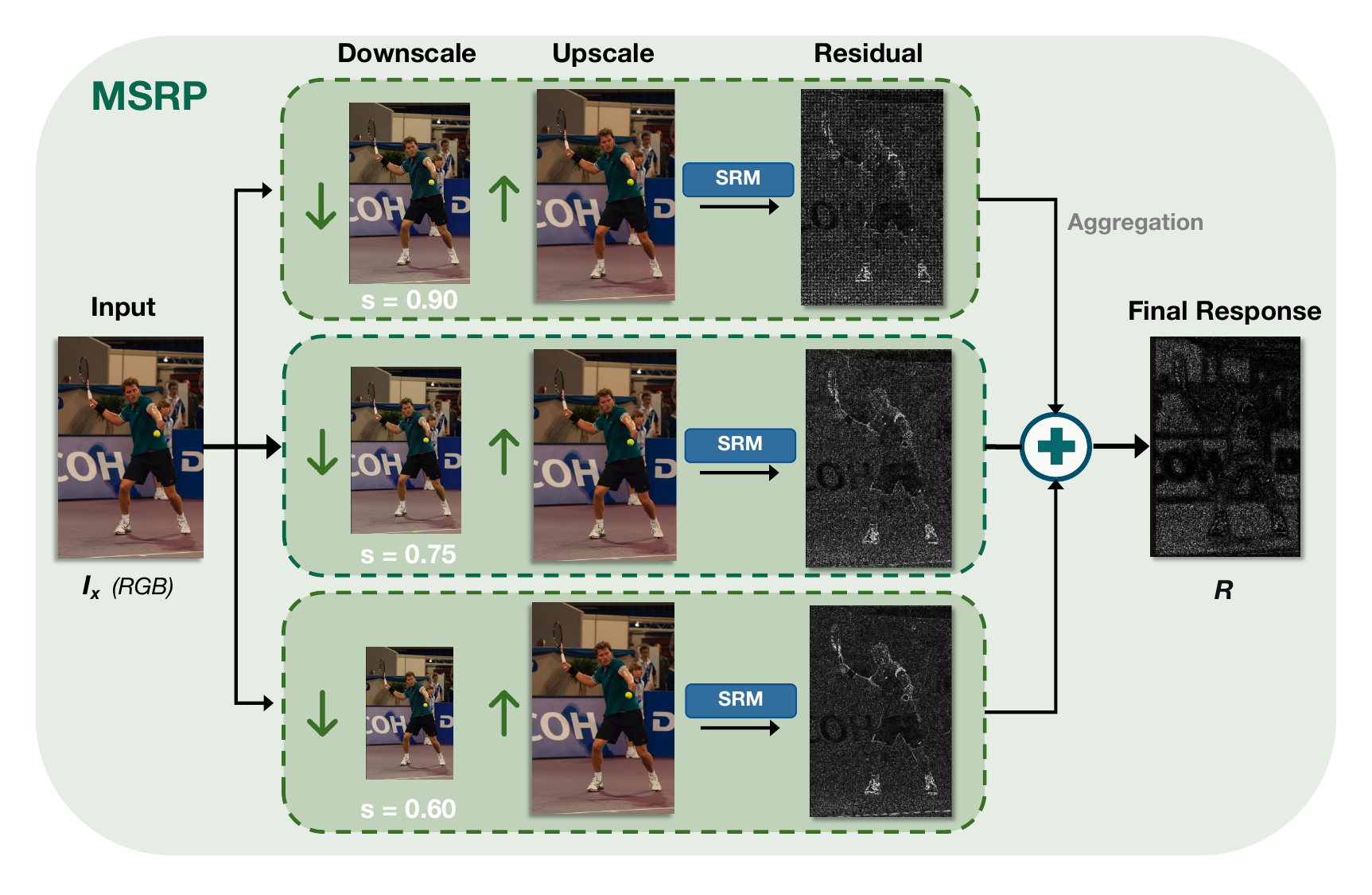}
\caption{Construction of the MSRP response map \(R\). The input image $x$ is resampled at three scales, and the SRM high-pass residuals between \(x\) and each \(x^{(s)}\) are averaged to form \(R\). MSRP has no learnable parameters.}
\label{fig:msrp-pipeline}
\end{figure*}

\begin{figure*}[htbp]
\centering
\includegraphics[width=0.88\textwidth]{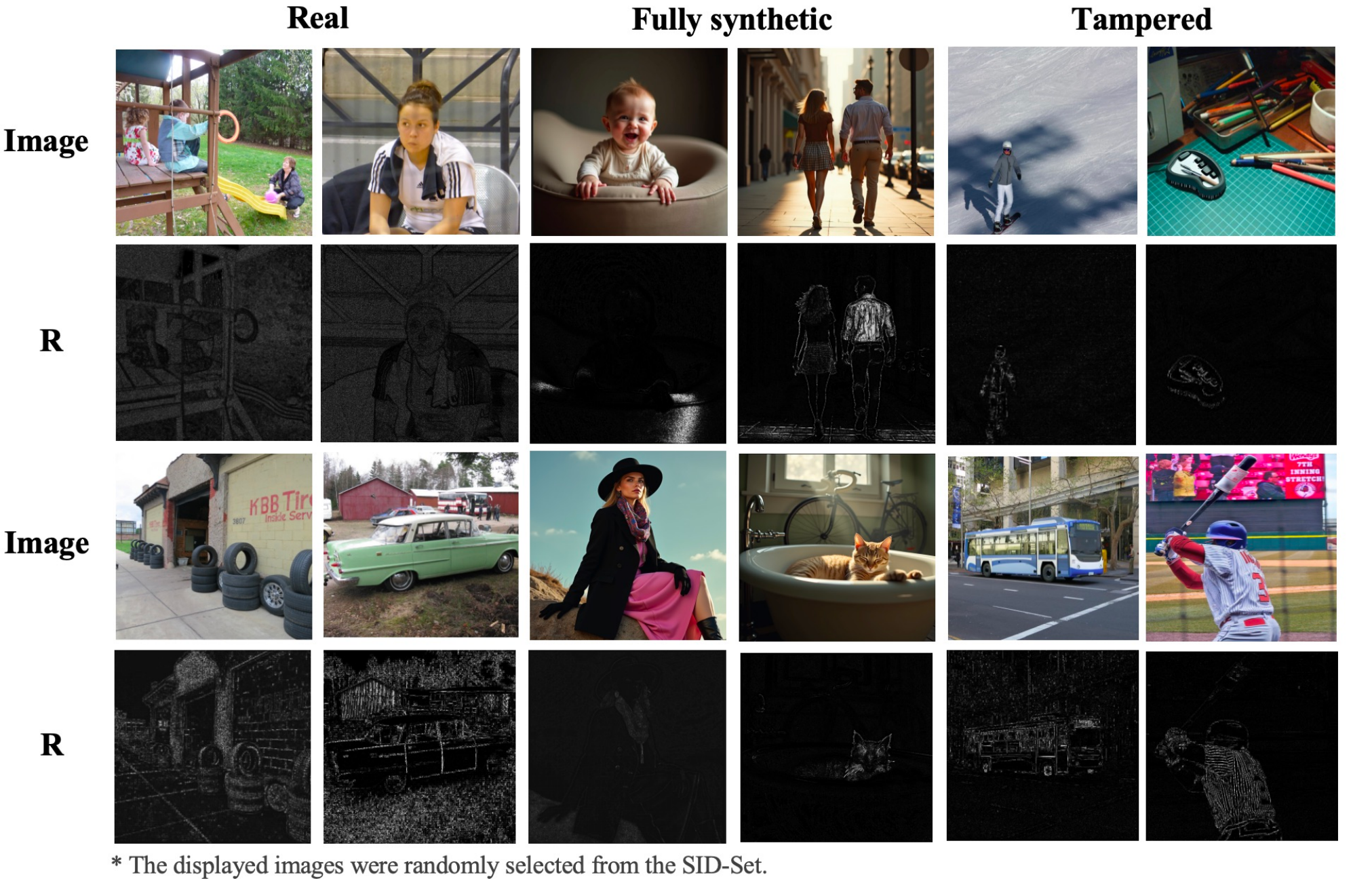}
\caption{Representative MSRP responses from the SID-Set. Each image is paired with its response map $R$. The examples illustrate broad responses in real images, concentrated responses near structural transitions in fully synthetic images, and local enrichment around manipulated content in tampered images.}
\label{fig:msrp_panels}
\end{figure*}

\paragraph{Visualizing the multi-scale construction.}
Figure~\ref{fig:msrp-pipeline} shows how the three branches probe local high-frequency structure at different resampling strengths. Following the construction in {Eq.~\ref{eq:msrp}} of the main text , \(R\) is computed by averaging the SRM high-frequency residuals between \(x\) and its restored views \(x^{(s)}\) over the scale set \(\mathcal{S}\). As Figure~\ref{fig:msrp-pipeline} summarizes: Each residual map highlights where local high-frequency responses change after a resample, a bright region therefore indicates sensitivity to resampling, which may arise from natural texture, boundaries, post-processing, or manipulation-related structure. The final map averages these responses across all three scales, so each branch contributes equally and no single scale serves as the base response. \textbf{MSRP} has no learnable parameters, and the scale set $\mathcal{S}=\{0.60,0.75,0.90\}$ is fixed across datasets. The resulting $R$ is supplied to the learned fusion module as a candidate low-level prior.

\paragraph{Response patterns across the three classes.}
Figure~\ref{fig:msrp_panels} highlights a central observation: real images produce broad and pronounced responses, whereas the two fake classes each exhibit distinct spatial patterns.

\begin{itemize}
    \item \textbf{Real images} produce strong and spatially broad responses. These responses extend across natural textures and object boundaries throughout the image.
    \item \textbf{Fully synthetic images} produce weaker but more concentrated responses. Strong responses cluster near structural boundaries and texture transitions, while large areas exhibit low response intensity.
    \item \textbf{Tampered images} exhibit localized response enrichment around manipulated content. These responses highlight manipulation boundaries against the broader responses of the natural background.
\end{itemize}

These qualitative examples suggest that the class-dependent signal lies in the spatial distribution of $R$, but do not by themselves establish a dataset-level trend.

\begin{table}[htbp]
\centering
\small
\setlength{\tabcolsep}{10pt}
\begin{tabular}{ccc}
\toprule
Class & Mean $R$ & Top-1\% energy \\
\midrule
Real      & 0.0555 & 0.0750 \\
Fully Synthetic & 0.0279 & 0.1131 \\
Tampered  & 0.0325 & 0.0753 \\
\bottomrule
\end{tabular}
\caption{Class-conditional MSRP response statistics on the SID-Set. Mean \(R\) is the average response intensity over all pixels. Top-1\% denotes the fraction of total response energy carried by the brightest $1\%$ of pixels. Values are averaged within each class.}
\label{tab:msrp_stats}
\end{table}

\paragraph{Quantitative context for the visual observations.}
As reported in Table~\ref{tab:msrp_stats}, the class averages are consistent with the visual observations and reveal a distinct class-dependent response pattern: real images have the highest mean response (\(0.0555\)), fully synthetic images (\(0.0279\)) have the greatest response concentration, and tampered images sit between them with an intermediate mean (\(0.0325\)) but a concentration close to that of real images. This ordering reflects the fact that $R$ measures sensitivity to resampling rather than the likelihood of forgery. The strong, spatially broad responses in real images may arise because acquisition noise, fine texture, and compression residuals are altered by resampling. Fully synthetic images often exhibit smoother, more regular local structure associated with generative upsampling and interpolation, which yields a lower mean response while concentrating it at structural and texture transitions. The brightest \(1\%\) of pixels in these images carry \(0.1131\) of the response energy, compared with \(0.0750\) for real images. This higher concentration indicates that their scale-sensitive high-frequency changes are localized near structural and texture transitions, even though their overall response magnitude is lower. Tampered images combine a natural background with locally generated content, consistent with their intermediate mean and the local enrichment in Figure~\ref{fig:msrp_panels}.

These statistics show that $R$ contains class-dependent spatial structure, while neither magnitude nor concentration alone is decisive. Since semantic boundaries, natural texture, and post-processing can also affect the response, \textbf{MSRP} serves as a general low-level candidate artifact prior, and the subsequent learned fusion determines which responses contribute to the evidence map $E$.

\section{Details of the Evidence Regularizer}
\label{app:regularizer}

The grouped term $\lambda_e\mathcal{L}_{\mathrm{evi}}$ complements classification supervision with class-conditioned spatial guidance. In three-class forgery detection, tampered images have manipulation masks, while real and fully synthetic images typically have only image-level labels, as in SID-Set~\cite{huang2025sida}. The regularizer addresses this imbalance by providing each class with a spatial constraint.

Let $[a]_+ = \max(a,0)$. For fully synthetic images, $\mathcal{H}$ and $\mathcal{L}$ contain the top and bottom $20\%$ of patches ranked by $R$. For tampered images, $A_i \in [0,1]$ is the manipulated pixel fraction in patch $i$, obtained by average-pooling the ground-truth mask. We use $A$ for this mask and reserve $M$ for the \textbf{EMV} visibility mask (Appendix~\ref{app:emv}). The region-weighted mean is $\operatorname{Mean}_{w}(v) = \sum_i w_i v_i / \sum_i w_i$, where $w=A$ or $1-A$. The three class-conditioned terms are
\begin{equation}
\begin{aligned}
\mathcal{L}_{\mathrm{real}}
&= \operatorname{Mean}(E)
 + 0.5\operatorname{Mean}_{\mathrm{top}\,10\%}(E),\\
\mathcal{L}_{\mathrm{syn}}
&= \bigl[
  0.1 -
  \bigl(
    \operatorname{Mean}_{\mathcal{H}}(E)
    - \operatorname{Mean}_{\mathcal{L}}(E)
  \bigr)
\bigr]_+,\\
\mathcal{L}_{\mathrm{tam}}
&= \operatorname{Mean}_{A}
   \bigl([\tau_{\mathrm{in}}-E]_+\bigr)\\
&\quad
 + \lambda_{\mathrm{out}}
   \operatorname{Mean}_{1-A}
   \bigl([E-\tau_{\mathrm{out}}]_+\bigr).
\end{aligned}
\label{eq:l-evi}
\end{equation}
\textbf{These terms provide coarse spatial guidance rather than fixed evidence targets. }$\operatorname{Mean}_{\mathrm{top}\,10\%}(E)$ averages the highest $10\%$ of evidence scores, so $\mathcal{L}_{\mathrm{real}}$ penalizes both average activation and isolated high-scoring patches. The margin in $\mathcal{L}_{\mathrm{syn}}$ discourages uniformly dark or bright maps while letting $E$ depart from the \textbf{MSRP} ordering, which is only a candidate prior. In $\mathcal{L}_{\mathrm{tam}}$, the mask provides soft support rather than a dense segmentation target, so evidence may concentrate on boundaries.

We set $(\tau_{\mathrm{in}},\tau_{\mathrm{out}}) = (0.4,0.12)$ and $\lambda_{\mathrm{out}}=0.7$. The real and tampered terms use weights of $0.5$ and $0.6$, respectively; the synthetic term uses $0.05$ during warm-up and $0.10$ thereafter. In short, $\mathcal{L}_{\mathrm{evi}}$ first shapes $E$, and \textbf{EMV} then tests and refines it toward decision-relevant evidence.

\section{EMV: Mask, Branches, Loss and Schedule}
\label{app:emv}

This section specifies the \textbf{EMV} mask, intervention branches, loss weights, uniform-map bound, and training schedule.
For $E=\{e_i\}_{i=1}^{N}$ with mean $\bar e=\frac{1}{N}\sum_{i=1}^{N}e_i$, we define the soft mask
\begin{equation}
M_i=\sigma\!\left(\frac{e_i-\bar e}{\tau}\right),
\label{eq:emv-mask}
\end{equation}
where $\tau$ sets its sharpness. Centering ties $M_i$ to how each score differs from the image mean.

Using shared evidence-weighted pooling and a shared classification head, we compute
\begin{equation}
\begin{aligned}
F_{\mathrm{det}}^{\mathrm{ori}}
&=\frac{\sum_i(1+e_i)f_i^{\mathrm{enc}}}
        {\sum_i(1+e_i)},\\
F_{\mathrm{det}}^{\mathrm{keep}}
&=\frac{\sum_i(1+e_i)M_i f_i^{\mathrm{enc}}}
        {\sum_i(1+e_i)},\\
F_{\mathrm{det}}^{\mathrm{drop}}
&=\frac{\sum_i(1+e_i)(1-M_i)f_i^{\mathrm{enc}}}
        {\sum_i(1+e_i)}.
\end{aligned}
\label{eq:emv-branches}
\end{equation}
For each branch $b\in\{\mathrm{ori},\mathrm{keep},\mathrm{drop}\}$, $P_{\mathrm{fake}}^b$ sums its synthetic and tampered probabilities.

Adding the term weights omitted in the main text gives
\begin{equation}
\begin{aligned}
\mathcal{L}_{\mathrm{emv}}
={}&\lambda_{\mathrm{keep}}
  \left(-\log P_{\mathrm{fake}}^{\mathrm{keep}}\right)\\
&+\lambda_{\mathrm{drop}}
  \max\!\left(0,\,
  m-\left(P_{\mathrm{fake}}^{\mathrm{ori}}
          -P_{\mathrm{fake}}^{\mathrm{drop}}\right)\right),
\end{aligned}
\label{eq:emv-full}
\end{equation}
with $(\lambda_{\mathrm{keep}},\lambda_{\mathrm{drop}})=(0.15,0.10)$ across both intervention phases. We apply \textbf{EMV} only to fake training examples that the original branch already predicts as fake, since misclassified examples lack a reliable baseline response. Before building the intervention branches, we detach $M$ so gradients do not flow through Eq.~\eqref{eq:emv-mask}, while $(1+e_i)$ stays differentiable.

When $E$ is uniform, $M_i=0.5$ and both intervention features reduce to $0.5F_{\mathrm{det}}^{\mathrm{ori}}$, so both branches give the same fake probability $u$. Write $p=P_{\mathrm{fake}}^{\mathrm{ori}}$. Splitting on $u\leq p-m$ and $u>p-m$, and using $-\log u\geq 1-u$ together with $\lambda_{\mathrm{keep}}\geq\lambda_{\mathrm{drop}}$, gives
\begin{equation}
\mathcal{L}_{\mathrm{emv}}
\geq
\min\!\left\{
-\lambda_{\mathrm{keep}}\log(1-m),\,
\lambda_{\mathrm{drop}}m
\right\}>0.
\label{eq:uniform-bound}
\end{equation}
A uniform map therefore has a strictly positive loss floor, though this bound on its own does not mean the full objective rules it out.

Table~\ref{tab:stage1-schedule} gives the schedule by phase. \textbf{EMV} stays off during head warm-up and evidence adaptation until the classifier and initial evidence map stabilize, then turns on for soft and hard intervention.

\section{Implementation and Training Details}
\label{app:implementation}

This section covers optimization settings and stage details omitted from the main text. All runs use seeds 418, 2026, and 3407 on two NVIDIA RTX 3090 GPUs.

\subsection{Stage 1: Detector and Evidence Learning}

\begin{table*}[!t]
\centering
\small
\setlength{\tabcolsep}{5pt}
\begin{tabular}{cccccc}
\toprule
Phase  & Active LoRA blocks & EMV &
$\tau$ & $m$ &
$(\lambda_{\rm keep},\lambda_{\rm drop})$ \\
\midrule
Head warm-up
& None & Off
& $0.50$ & $0.10$ & $(0,0)$ \\
Evidence adaptation
& 18--23 & Off
& $0.50$ & $0.10$ & $(0,0)$ \\
Soft intervention
& 12--23 & On
& $0.50\rightarrow0.35$ & $0.10$ & $(0.15,0.10)$ \\
Hard intervention
& 6--23 & On
& $0.35\rightarrow0.30$ & $0.10\rightarrow0.15$
& $(0.15,0.10)$ \\
\bottomrule
\end{tabular}
\caption{Progressive Stage-1 training schedule. EMV is introduced only after the detector and the initial evidence map have stabilized.}
\label{tab:stage1-schedule}
\end{table*}

We square-crop each image along its shorter side, resize it to $512\times512$, and encode it on a $32\times32$ patch grid using a frozen DINOv3 ViT-L/16 with LoRA adapters on selected query and value projections. \textbf{MSRP} response maps are precomputed at $512\times512$ over scales $\{0.60,0.75,0.90\}$ and aligned with the inputs. No additional stochastic image degradation augmentation is used in the main SID-Set experiment.

Stage 1 uses AdamW, bfloat16, a global batch size of 16, and gradient clipping at 1.0. The evidence fusion module, \textbf{MSRP} tokenizer, evidence head, and detection head use a learning rate of $3\times10^{-4}$ and weight decay of 0.05. Positional embeddings and learned evidence-path scalars use $1\times10^{-4}$ with zero weight decay, while LoRA parameters use group-specific learning rates and weight decay of 0.01. Learning rates use linear warm-up followed by cosine decay to 1\% of their initial values. Table~\ref{tab:stage1-schedule} gives the schedule by phase. The first two phases establish the detection boundary and the initial evidence distribution without \textbf{EMV}. The soft-intervention phase then activates \textbf{EMV} with a smooth visibility mask, after which the hard-intervention phase sharpens the mask and raises the margin.

\subsection{Stage 2: Evidence-Guided Explanation}

After Stage 1, we freeze the detector and cache its features, predicted classes, and evidence maps on the training set, skipping the Qwen3-VL vision tower on raw images. Stage 2 runs for $5$ epochs: the first two train only the projectors with $\mathcal{L}_{\mathrm{sft}}$, and the last three jointly train the projectors and rank-16 Qwen3-VL-8B LoRA adapters with $\mathcal{L}_{\mathrm{sft}} + \lambda_{\mathrm{ec}}\mathcal{L}_{\mathrm{ec}}$ ($\lambda_{\mathrm{ec}}=0.2$, margin $m=0.1$). The base Qwen3-VL weights stay frozen, with LoRA scaling at 32 and dropout at 0.05.

Stage 2 uses AdamW and bfloat16 with a micro-batch size of 1, an effective batch size of 8, and a projector learning rate of $1\times10^{-4}$.

\section{Additional Detection Results}

\subsection{Backbone Control and Model Size}
\begin{table*}[t]
\centering
\small
\setlength{\tabcolsep}{12pt}
\begin{tabular}{ccccc}
\toprule
Method & Acc (\%) & F1 (\%) & Params (M) & ms/img \\
\midrule
SIDA & $93.6{\pm}2.22$ & $93.5{\pm}2.39$ & 12.99 & $201.69{\pm}24.02$ \\
CLIP (LoRA) & $85.4{\pm}0.11$ & $86.0{\pm}0.12$ & 1.28 & $29.04{\pm}1.68$ \\
EviDet (CLIP) & \underline{$98.1{\pm}0.09$} & \underline{$98.1{\pm}0.09$} & 5.81 & $41.36{\pm}1.95$ \\
EviDet (DINOv3) & \bm{$99.4{\pm}0.09$} & \bm{$99.4{\pm}0.09$} & 6.60 & $45.03{\pm}2.41$ \\
\bottomrule
\end{tabular}
\caption{Backbone control and cost on SID-Set (mean $\pm$ standard deviation over three seeds). Params: trainable parameters of the detector; ms/img: detector inference time per image.}
\label{tab:backbone}
\end{table*}

To separate the method from the backbone, we train EviDet with a CLIP-L/14 backbone at 
$224\times224$ under the same settings (Table~\ref{tab:backbone}). CLIP with LoRA alone reaches 85.4 accuracy, while our fine-tuned CLIP reaches 98.1, a gain of 12.7 points with only 4.53M more trainable parameters and 12.32 ms more per image. The larger gain on CLIP shows the method does not rely on a strong backbone, and the extra 1.3 points on DINOv3 shows backbone quality still helps marginally. EviDet (DINOv3) also outperforms SIDA by 5.8 points while using 6.39M fewer trainable parameters and 156.66 ms less per image.

\subsection{Robustness to Common Perturbations}
\label{app:robustness}

\begin{figure*}[t]
\centering
\includegraphics[width=1.0\textwidth]{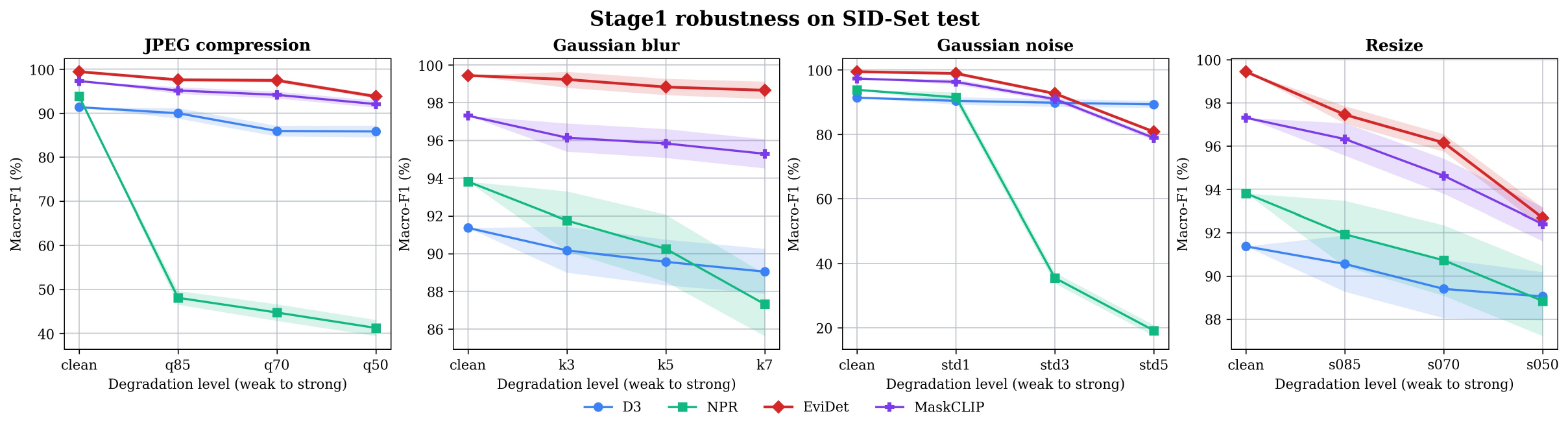}
\caption{Stage-1 robustness on the SID-Set test set under JPEG compression, Gaussian blur, Gaussian noise, and resizing. Markers and connecting lines report the mean Macro-F1 over three evaluation seeds, while the shaded regions span the minimum and maximum values. Each method uses one fixed checkpoint across all perturbations.}
\label{fig:robustness}
\end{figure*}

We evaluate three-class detection on SID-Set under four image perturbations.

\paragraph{Protocol.}
Each method uses one fixed checkpoint trained on the clean SID-Set training split. We apply one perturbation at a time to the fixed test split and recompute \textbf{MSRP} maps from the perturbed inputs. Following SPAI~\cite{karageorgiou2025any}, we use JPEG quality levels $\{85,70,50\}$, Gaussian blur kernels of size $k\times k$ with $k\in\{3,5,7\}$, Gaussian noise with standard deviation $\sigma\in\{1,3,5\}$, and resizing scales $\{0.85,0.70,0.50\}$. Macro-F1 is the metric throughout.

\paragraph{Results.}
EviDet ranks first in 11 of the 12 settings, averaging 95.35\% Macro-F1 against 93.17\% for MaskCLIP, 89.09\% for D$^3$, and 68.41\% for NPR. At the strongest JPEG, blur, and resizing levels it retains 93.83\%, 98.66\%, and 92.67\%. The only non-leading case is strong Gaussian noise, where EviDet reaches 80.78\% and D$^3$ reaches 89.30\%.

Seed variation separates repeatability from average robustness. Across the 12 settings, EviDet has an average min--max band of 0.87 points and a maximum of 0.95 points, compared with 1.55, 2.43, and 3.34 for MaskCLIP, D$^3$, and NPR. EviDet combines the highest average robustness with the lowest sensitivity to the perturbation realization.

Blur is where EviDet is most stable, remaining leading even at $7\times7$, and it stays ahead of MaskCLIP by 1.74 and 0.28 points under the strongest JPEG and resizing. Compression, blur, and downsampling weaken or reorganize the high-frequency responses \textbf{MSRP} relies on, and the visual features compensate for the weakened low-level evidence. NPR depends more directly on low-level pixel statistics and degrades accordingly.

Gaussian noise is the main limitation. EviDet retains 98.91\% at $\sigma=1$, drops to 92.65\% at $\sigma=3$, and to 80.78\% at $\sigma=5$. Additive noise injects high-frequency energy across the image and obscures the structured resampling residuals \textbf{MSRP} isolates. D$^3$ is less affected and overtakes only at $\sigma=5$.

\subsection{Cross-Domain Tampering Detection on MMTD-Set}
\label{app:mmtd-detection}

\begin{table*}[!t]
\centering
\small
\setlength{\tabcolsep}{7pt}
\begin{tabular}{c|cccccccccc|cc|cc}
\toprule
\multirow{3}{*}{Method} &
\multicolumn{10}{c|}{\textbf{PhotoShop}} &
\multicolumn{2}{c|}{\multirow{2}{*}{\textbf{DeepFake}}} &
\multicolumn{2}{c}{\multirow{2}{*}{\textbf{AIGC-Editing}}} \\
&
\multicolumn{2}{c}{CASIA1+} &
\multicolumn{2}{c}{IMD2020} &
\multicolumn{2}{c}{Columbia} &
\multicolumn{2}{c}{Coverage} &
\multicolumn{2}{c|}{DSO} &
& &
& \\
\cmidrule(lr){2-3} \cmidrule(lr){4-5} \cmidrule(lr){6-7} \cmidrule(lr){8-9} \cmidrule(lr){10-11} \cmidrule(lr){12-13} \cmidrule(lr){14-15}
& ACC & F1
& ACC & F1
& ACC & F1
& ACC & F1
& ACC & F1
& ACC & F1
& ACC & F1 \\
\midrule
SPAN & 0.60 & 0.44 & 0.70 & 0.81 & 0.87 & 0.93 & 0.24 & 0.39 & 0.35 & 0.52 & 0.78 & 0.78 & 0.47 & 0.05 \\
ManTraNet & 0.52 & 0.68 & 0.75 & 0.85 & \underline{0.95} & \underline{0.97} & 0.95 & \underline{0.97} & 0.90 & 0.95 & 0.50 & 0.67 & 0.50 & 0.67 \\
HiFi-Net & 0.46 & 0.44 & 0.62 & 0.75 & 0.68 & 0.81 & 0.34 & 0.51 & 0.96 & \underline{0.98} & 0.56 & 0.61 & 0.49 & 0.42 \\
PSCC-Net & 0.90 & 0.89 & 0.67 & 0.78 & 0.78 & 0.87 & 0.84 & 0.91 & 0.66 & 0.80 & 0.48 & 0.58 & 0.49 & 0.65 \\
CAT-Net & 0.88 & 0.87 & 0.68 & 0.79 & 0.89 & 0.94 & 0.23 & 0.37 & 0.86 & 0.92 & 0.85 & 0.84 & 0.82 & 0.81 \\
MVSS-Net & 0.62 & 0.76 & 0.75 & 0.85 & 0.94 & \underline{0.97} & 0.65 & 0.79 & 0.96 & \underline{0.98} & 0.84 & 0.91 & 0.44 & 0.24 \\
FakeShield & \underline{0.95} & \underline{0.95} & \underline{0.83} & \underline{0.90} & \textbf{0.98} & \textbf{0.99} & \textbf{0.97} & \textbf{0.98} & \underline{0.97} & \underline{0.98} & \underline{0.93} & \underline{0.93} & \underline{0.98} & \underline{0.99} \\
\midrule
EviDet (ours) & \textbf{0.99} & \textbf{0.98} & \textbf{0.94} & \textbf{0.95} & \textbf{0.98} & \textbf{0.99} & \underline{0.96} & \textbf{0.98} & \textbf{0.98} & \textbf{0.99} & \textbf{0.95} & \textbf{0.97} & \textbf{1.00} & \textbf{1.00} \\
\bottomrule
\end{tabular}
\caption{Cross-domain binary tampering detection under the official MMTD-Set protocol. Results for other methods are reported by~\citet{xu2025fakeshield}. Best and second-best results are shown in \textbf{bold} and \underline{underline}, respectively.}
\label{tab:MMTD-Set}
\end{table*}

The OpenSDID experiment evaluates cross-generator generalization but does not cover the transfer of locally manipulated evidence across editing domains. We therefore conduct binary real-versus-tampered detection on MMTD-Set~\cite{xu2025fakeshield}, whose fake class consists entirely of locally manipulated images. Following the official cross-domain protocol, EviDet is trained on the designated split and evaluated directly on seven test sets, without target-domain adaptation. Thus, this experiment measures generalization across unseen manipulation sources and editing pipelines, rather than zero-shot transfer from SID-Set.

As shown in Table~\ref{tab:MMTD-Set}, EviDet achieves the best or tied-best result in 13 of the 14 domain--metric combinations. Averaged over the seven test sets, it obtains 0.971 accuracy and 0.980 F1, exceeding the strongest baseline, FakeShield, by 2.7 and 2.0 percentage points, respectively. The gains remain consistent across conventional Photoshop manipulations, DeepFake, and AIGC-based editing: EviDet reaches 1.00/1.00 on AIGC-Editing and 0.95/0.97 on DeepFake, while its lowest result across all domains remains 0.94 accuracy and 0.95 F1 on IMD2020.

This experiment complements the fully synthetic generalization results on OpenSDID. Since every fake sample in MMTD-Set is locally manipulated, the gains do not stem from fully generated-image artifacts, but from the model's generalization to local edits.

\subsection{Transfer to an Autoregressive Generator}
\label{app:var}

\begin{table}[htbp]
\centering
\small
\setlength{\tabcolsep}{12pt}
\begin{tabular}{ccc}
\toprule
Method & Acc$\uparrow$ & F1$\uparrow$ \\
\midrule
MaskCLIP & 62.37 & 56.63 \\
\textbf{EviDet (ours)} & \textbf{82.89} & \textbf{82.84} \\
\bottomrule
\end{tabular}
\caption{Binary detection on the 12,000-image VAR subset of ARForensics (\%).}
\label{tab:var}
\end{table}

\paragraph{Protocol.}
To evaluate transfer beyond diffusion models, we test EviDet and MaskCLIP on the VAR subset of the ARForensics test set~\cite{zhang2025d3qe}. The subset contains 12,000 images, with 6,000 real and 6,000 generated by VAR~\cite{tian2024visual}. Both detectors are trained on SID-Set and evaluated on binary real-versus-fake detection without fine-tuning on ARForensics.

\paragraph{Results.}
As shown in Table~\ref{tab:var}, EviDet outperforms MaskCLIP by 20.52 and 26.21 percentage points in accuracy and F1, respectively. Unlike MaskCLIP, which relies primarily on semantic representations that degrade under generator shifts, EviDet leverages low-level residual cues that capture local pixel inconsistencies shared across different synthesis processes. Together with the results on unseen diffusion generators in OpenSDID, this suggests that EviDet can transfer across generation paradigms, while further optimization tailored to autoregressive artifacts is left for future work.

\subsection{Feature Space Visualization}
\begin{figure*}[htbp]
\centering
\includegraphics[width=0.8\textwidth]{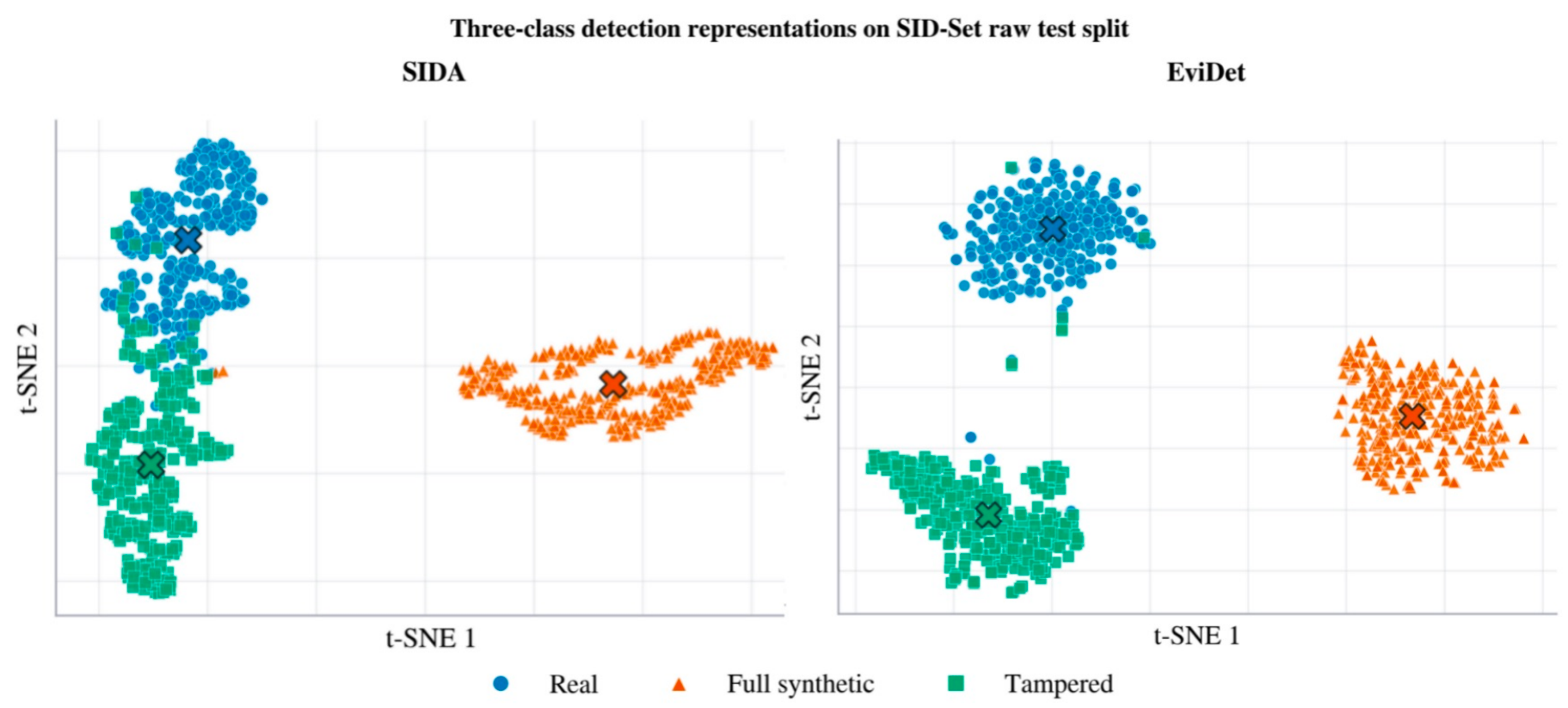}
\caption{t-SNE projections of pooled detection features on the SID-Set raw test split for SIDA (left) and EviDet (right). Colors and markers denote the three classes. Large crosses mark their projected centers. EviDet forms tighter clusters with less overlap between real and tampered samples.}
\label{fig:tsne}
\end{figure*}

To qualitatively assess the learned representations, we use t-SNE to project the pooled detection features of EviDet and SIDA~\cite{huang2025sida} on the SID-Set raw test split under the same three-class setting. For SIDA, we use the authors' officially released SIDA-7B weights. As shown in Figure~\ref{fig:tsne}, SIDA separates fully synthetic images, but its real and tampered features form elongated, overlapping manifolds. In contrast, EviDet yields three tighter, better-separated clusters, with a marked reduction in real-tampered overlap. Although t-SNE provides only a qualitative view, this structure is consistent with a more class-discriminative representation.

\section{Additional Analysis of the Evidence Map}
\subsection{Fixed Budgets on a Detector Trained without EMV}
\label{app:fixed-budget}

\begin{table}[htbp]
\centering
\small
\setlength{\tabcolsep}{12pt}
\begin{tabular}{cccc}
\toprule
Ranking & Budget & Suff. $\downarrow$ & Comp. $\uparrow$ \\
\midrule
\multirow{3}{*}{$E$} & 5\%  & 0.765  & 0.032 \\
                     & 20\% & 0.553  & 0.404 \\
                     & 50\% & 0.077  & 0.710 \\
\midrule
\multirow{3}{*}{Random} & 5\%  & 0.020  & -0.003 \\
                        & 20\% & -0.053 & -0.012 \\
                        & 50\% & 0.068  & 0.054 \\
\bottomrule
\end{tabular}
\caption{Fixed-budget intervention analysis on a detector trained without EMV.}
\label{tab:fixed_budget}
\end{table}

The analysis in the main text uses the \textbf{EMV}-trained detector and sets each map's budget by the mask area of $E$, the setting that the \textbf{EMV} losses optimize. Table~\ref{tab:fixed_budget} repeats it on a detector trained without \textbf{EMV} at fixed budgets of 5\%, 20\%, and 50\% of the patches. The median mask area of $E$ is 19\% on tampered and 55\% on fully synthetic images. Removing the top-ranked patches lowers the fake probability by 0.032, 0.404, and 0.710, while removing the same number of random patches changes it by -0.003, -0.012, and 0.054. Keeping only the top 5\% or 20\% costs 0.765 and 0.553, far more than keeping random patches, and at 50\% the two costs are close.

\subsection{Per-Image Keep/Drop Audit}
\begin{table}[htbp]
\centering
\small
\setlength{\tabcolsep}{12pt}
\begin{tabular}{cccc}
\toprule
Class & Keep & Drop & Joint \\
\midrule
Fully Synthetic & 98.2 & 95.6 & 94.1 \\
Tampered & 97.3 & 94.4 & 92.9 \\
\bottomrule
\end{tabular}
\caption{Per-image keep/drop pass rates (\%) on the fake predictions of the SID-Set test split ($m=0.1$).}
\label{tab:pass_rate}
\end{table}

The same two interventions (keep/drop) in \textbf{EMV} can be rerun on any image after training. For each SID-Set test image predicted as fake, the keep test passes if the keep branch preserves the prediction, and the drop test passes if the drop branch lowers $P_{\rm fake}$ by at least $m=0.1$, joint requires both. Table~\ref{tab:pass_rate} reports the pass rates: 94.1\% of fully synthetic and 92.9\% of tampered predictions pass both. The joint rate is lower than either single rate because a small fraction of predictions satisfy only one of the two conditions. This result validates our design choice of not requiring \textbf{EMV} at inference: with \textbf{EMV} disabled, the keep and drop interventions still hold on individual images, and the vast majority of fake predictions pass both.

\subsection{Forgery Localization on SID-Set}
\label{app:localization}

\paragraph{Protocol.}
The main text evaluates $E$ primarily as a decision-support evidence map through keep/drop interventions. Here we provide a complementary spatial analysis on the tampered subset of SID-Set, where ground-truth manipulation masks are available. Following the evaluation protocol of SIDA~\cite{huang2025sida}, we upsample the $32\times32$ patch-level evidence map $E$ to the mask resolution and evaluate its alignment using AUC, F1, and IoU. 

\begin{table}[htbp]
\centering
\small
\setlength{\tabcolsep}{8.5pt}
\begin{tabular}{ccccc}
\toprule
Method & Venue & AUC$\uparrow$ & F1$\uparrow$ & IoU$\uparrow$ \\
\midrule
MVSS-Net & TPAMI'22 & 48.9 & 31.6 & 23.7 \\
HIFI-Net & CVPR'23 & 64.0 & 45.9 & 21.1 \\
PSCC-Net & TCSVT'22 & 82.1 & 71.3 & 35.7 \\
LISA-7B & CVPR'24 & 78.4 & 69.1 & 32.5 \\
SIDA-7B & CVPR'25 & 87.3 & \underline{73.9} & 43.8 \\
MaskCLIP & CVPR'25 & \underline{96.4} & 63.9 & \underline{54.5} \\
\midrule
\textbf{EviDet (ours)} & -- & \textbf{99.1} & \textbf{75.0} & \textbf{64.2} \\
\bottomrule
\end{tabular}
\caption{Spatial alignment with ground-truth manipulation masks on the tampered subset of SID-Set. Results for MVSS-Net, HIFI-Net, PSCC-Net, LISA-7B, and SIDA-7B are reported by~\citet{huang2025sida}; MaskCLIP is reproduced using its official implementation under the same protocol. Best and second-best results are shown in \textbf{bold} and \underline{underline}, respectively.}
\label{tab:localization}
\end{table}

\paragraph{Results and Discussion.}
$E$ achieves 99.1 AUC, 75.0 F1, and 64.2 IoU, the best on all three metrics (Table~\ref{tab:localization}). Against the strongest competing result per metric, EviDet improves AUC over MaskCLIP by 2.7 points, F1 over SIDA-7B by 1.1 points, and IoU over MaskCLIP by 9.7 points.

$E$ is trained with these masks as soft support (Appendix~\ref{app:regularizer}) and has no segmentation head, so the comparison shows how well a $32\times32$ evidence map localizes edits. This spatial alignment is what \textbf{EGE} conditions on when it grounds the explanation.

\section{Additional Analysis of the Explanation}

\begin{figure*}[!t]
\centering
\includegraphics[width=0.88\textwidth]{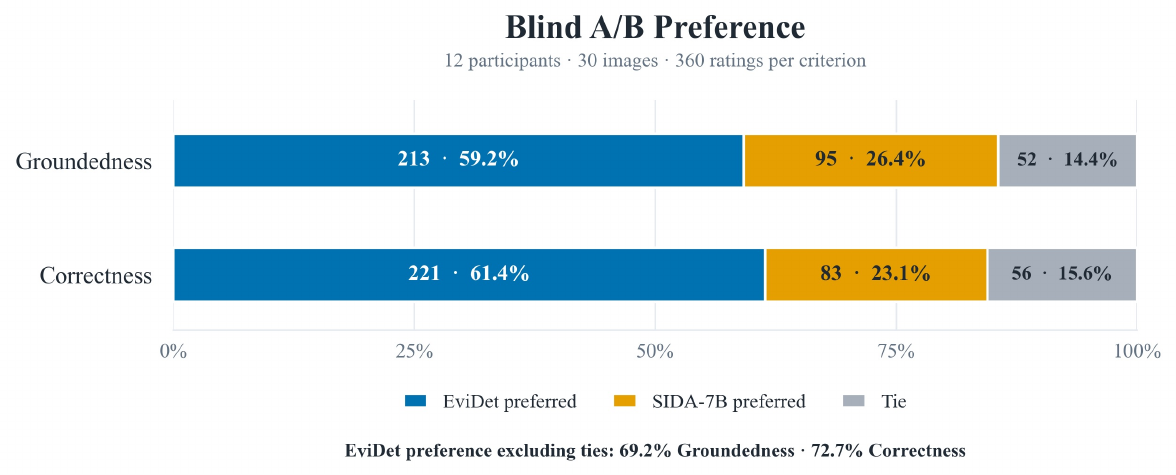}
\caption{Human preferences in a blind A/B comparison of EviDet and SIDA-7B. Bars show vote counts and shares of all 360 ratings per criterion. The reported EviDet preference rates exclude ties.}
\label{fig:human-preference}
\end{figure*}

This section provides additional analysis of the explanation module. We first ablate the evidence input and evidence-null contrastive decoding in \textbf{EGE}, then report a blind human A/B study that complements the automatic metrics.

\subsection{EGE Component Ablation}
Table~\ref{tab:ege_ablation} isolates the two core designs of the explanation generator: the evidence input $E$ and evidence-null contrastive decoding (CD). The first two rows measure the overall gain of conditioning on $E$. The second and last rows share the same evidence-conditioned checkpoint and differ only in decoding, which isolates the contribution of CD at inference. The third row removes $\mathcal{L}_{\rm ec}$ from training.

\begin{table}[htbp]
\centering
\small
\begin{tabular}{ccccc}
\toprule
Method & Evidence $E$ & CD & ROUGE-L$\uparrow$ & CSS$\uparrow$ \\
\midrule
EGE w/o $E$ & & & 41.09 & 77.34 \\
EGE w/ $E$ & \checkmark & & \textbf{42.25} & 78.31 \\
EGE w/o $\mathcal{L}_{\rm ec}$ & \checkmark & \checkmark & 42.02 & 78.97 \\
\textbf{EGE (ours)} & \checkmark & \checkmark & 42.11 & \textbf{82.23} \\
\bottomrule
\end{tabular}
\caption{Ablation of EGE on SID-Set. ROUGE-L and CSS are multiplied by 100. Except for the evidence-conditioning and CD switches shown in the table and the margin loss removed in the third row, all settings share the frozen detector, Qwen/LoRA configuration, prompt, and generation settings.}
\label{tab:ege_ablation}
\end{table}

As shown in Table~\ref{tab:ege_ablation}, conditioning on $E$ improves ROUGE-L from 41.09 to 42.25 and CSS from 77.34 to 78.31. Applying CD to the same checkpoint further raises CSS to 82.23, whereas removing $\mathcal{L}_{\rm ec}$ under the same decoding settings lowers it to 78.97. These results support combining evidence conditioning with evidence contrast to improve agreement with reference explanations.

\subsection{Human Evaluation}
\label{app:human-eval}

\paragraph{Protocol.}
To complement automatic explanation metrics with human judgments, we conducted a blind A/B comparison between EviDet and SIDA-7B on 30 SID-Set images, with 10 images sampled from each class (real, fully synthetic, and tampered). Twelve participants compared the anonymized explanations in randomized order using two criteria adapted from DeepfakeJudge~\cite{kuckreja2026deepfakejudge}: \textbf{Groundedness} (whether the cited visual cues are present and relevant to the conclusion) and \textbf{Correctness} (whether the explanation accurately describes the forensic traces). This yielded 360 ratings per criterion. Ties are included in the vote counts and excluded only when calculating preference rates.

\paragraph{Results.}
Excluding ties, non-tied votes favored EviDet in 69.2\% of Groundedness comparisons and 72.7\% of Correctness comparisons (Figure~\ref{fig:human-preference}). By anchoring text generation to the shared evidence map $E$, EviDet avoids ungrounded descriptions and aligns its explanations more closely with visible forensic traces. Together with the automatic metrics on TriDF, these human judgments confirm that EviDet produces more reliable and faithful explanations.

\section{Detection Component Ablations}
\label{app:detector-ablations}

\begin{table*}[htbp]
\centering
\small
\setlength{\tabcolsep}{7pt}
\begin{tabular}{cccccc}
\toprule
Method & $(\lambda_{\rm keep},\lambda_{\rm drop})$ & Acc$\uparrow$ & Macro-F1$\uparrow$ & $\Delta P_{\rm keep}\downarrow$ & $\Delta P_{\rm drop}\uparrow$ \\
\midrule
w/o EMV & $(0,0)$ & 92.6 & 92.5 & -- & -- \\
keep only & $(0.15,0)$ & 95.5 & 94.9 & 0.0105 & 0.1310 \\
drop only & $(0,0.10)$ & 95.7 & 95.3 & -0.0106 & 0.6821 \\
balanced & $(0.10,0.10)$ & 98.7 & 98.9 & 0.0705 & 0.7490 \\
asymmetric (ours) & $(0.15,0.10)$ & \textbf{99.4} & \textbf{99.4} & 0.0652 & 0.7879 \\
\bottomrule
\end{tabular}
\caption{Sensitivity to the EMV keep/drop weights on SID-Set. Weights denote the final-stage setting. Acc and Macro-F1 are reported in percent. $\Delta P_{\rm keep}$ and $\Delta P_{\rm drop}$ denote the decrease in fake probability after retaining or removing the selected patches, respectively, and are reported in probability units.}
\label{tab:emv_weights}
\end{table*}

\begin{table*}[htbp]
\centering
\small
\setlength{\tabcolsep}{8.5pt}
\begin{tabular}{ccccc}
\toprule
Setting & Cross-scale set & Acc$\uparrow$ & Macro-F1$\uparrow$ \\
\midrule
Single-0.60 & $\{0.60\}$ & 94.7 & 94.7 \\
Single-0.75 & $\{0.75\}$ & 95.2 & 95.3 \\
Single-0.90 & $\{0.90\}$ & 94.8 & 94.9 \\
Dual-0.60/0.90 & $\{0.60,0.90\}$ & 97.6 & 97.8 \\
Multi-scale (ours) & $\{0.60,0.75,0.90\}$ & \textbf{99.4} & \textbf{99.4} \\
\bottomrule
\end{tabular}
\caption{Sensitivity of EviDet to the MSRP residual scales on SID-Set.}
\label{tab:msrp_scales}
\end{table*}

\subsection{EMV Keep/Drop Weights}
Table~\ref{tab:emv_weights} shows that both \textbf{EMV} objectives matter and that the asymmetric configuration outperforms the symmetric one.

Relative to the model without \textbf{EMV}, combining the two intervention terms improves accuracy from 92.6 to 99.4 and Macro-F1 from 92.5 to 99.4. Either term alone provides only a moderate gain: the keep objective encourages the selected patches to remain sufficient for the fake prediction, whereas the drop objective makes their removal reduce the fake response. Their complementarity is also visible in the intervention statistics. Drop-only produces a large $\Delta P_{\rm drop}$ (0.6821), while keep-only yields little response change after the selected evidence is removed ($\Delta P_{\rm drop}=0.1310$). The asymmetric setting achieves the best joint operating point, reducing $\Delta P_{\rm keep}$ from 0.0705 to 0.0652 and increasing $\Delta P_{\rm drop}$ from 0.7490 to 0.7879 over the balanced setting. This directly supports the main-text conclusion that sufficiency and comprehensiveness must be optimized together.

\subsection{MSRP Scale Sensitivity}
Table~\ref{tab:msrp_scales} presents the ablation study metrics for the \textbf{MSRP} module at different scales. Among the single-scale variants, the $0.75$ residual performs best, reaching 95.2 accuracy and 95.3 Macro-F1, but every single scale remains substantially below the multi-scale design. Combining two scales raises the results to 97.6/97.8, and the complete set $\{0.60,0.75,0.90\}$ further reaches 99.4/99.4. The gain of 4.2 accuracy and 4.1 Macro-F1 over the best single-scale setting is consistent with the motivation in the main text: \textbf{MSRP} benefits from residual patterns that persist across resampling scales rather than relying on one scale-specific artifact signature.

\section{Additional Qualitative Results}
\label{app:qualitative}

\begin{figure*}[!t]
\centering
\includegraphics[width=0.9\textwidth]{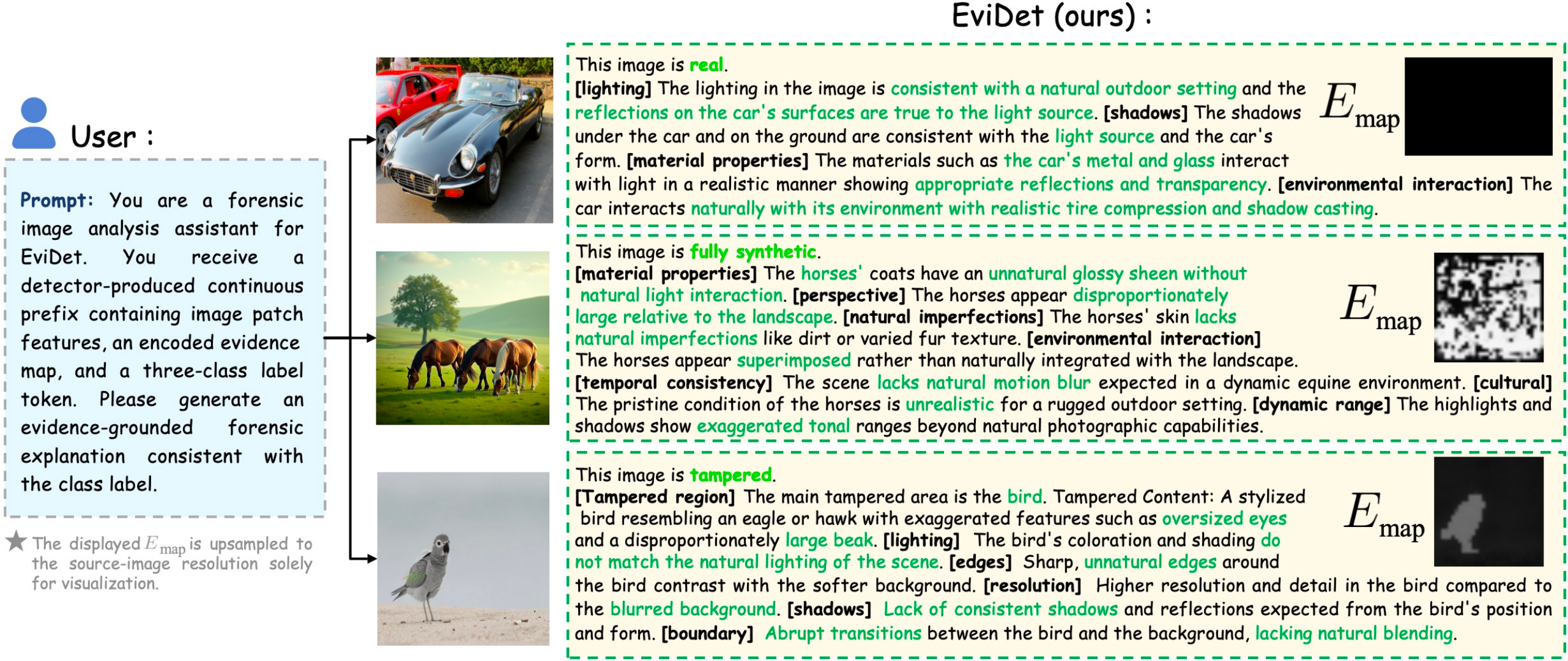}
\caption{Successful qualitative cases across the three EviDet classes. Green text highlights evidence-specific findings grounded in $E$: near-null evidence accompanies the real example, distributed evidence supports scene-wide synthetic artifacts, and localized evidence anchors the tampered explanation to the manipulated region.}
\label{fig:successful_cases}
\end{figure*}

\begin{figure}[htbp]
\centering
\includegraphics[width=\columnwidth]{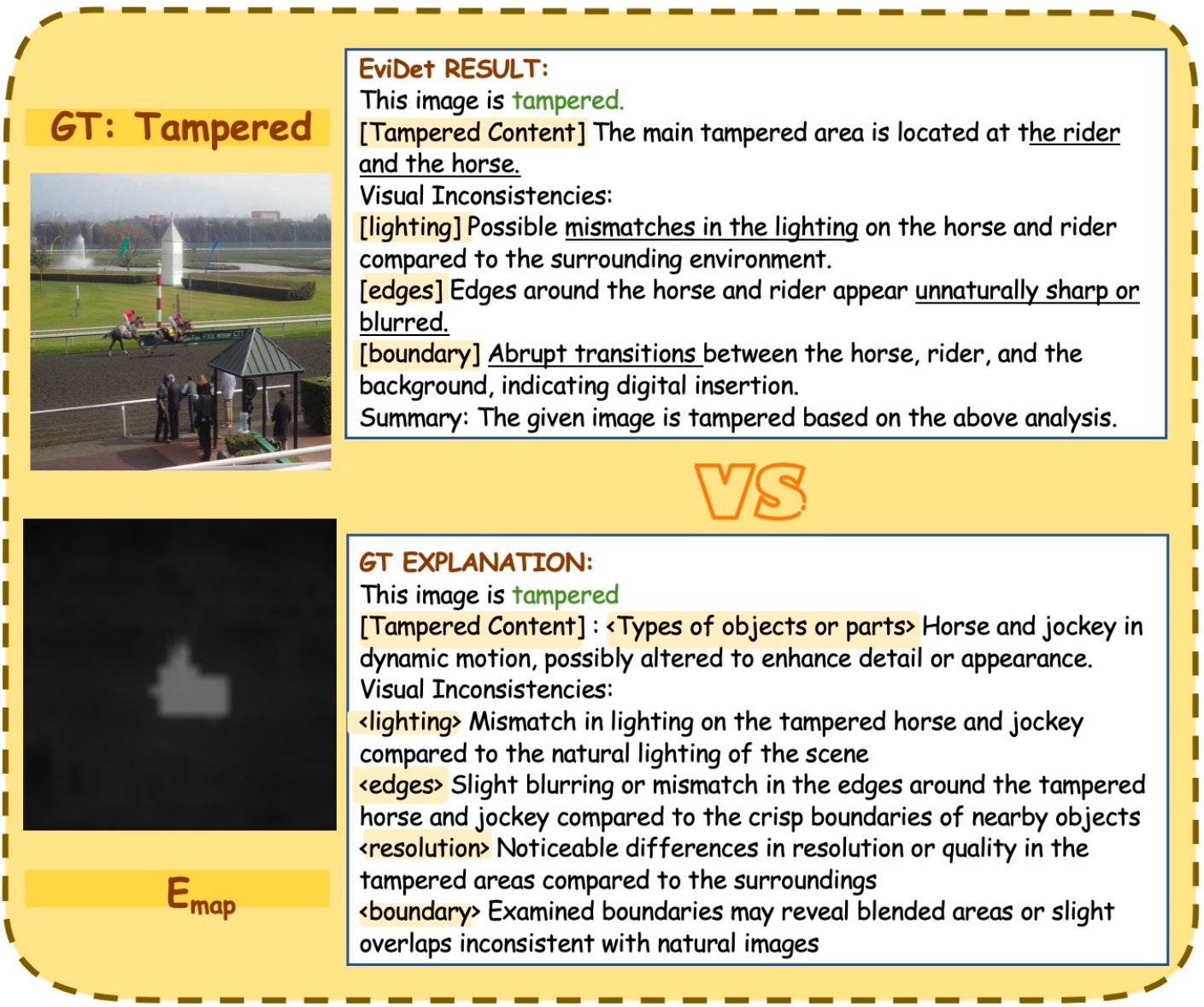}
\caption{Comparison of Contrastive Decoding (CD) with the ground-truth (GT) explanation. Both outputs correctly identify the image as tampered and agree on the core evidence, while CD omits weaker details. This illustrates the trade-off between conciseness and completeness.}
\label{fig:cd_tradeoff}
\end{figure}

\begin{figure*}[t]
\centering
\includegraphics[width=0.88\textwidth]{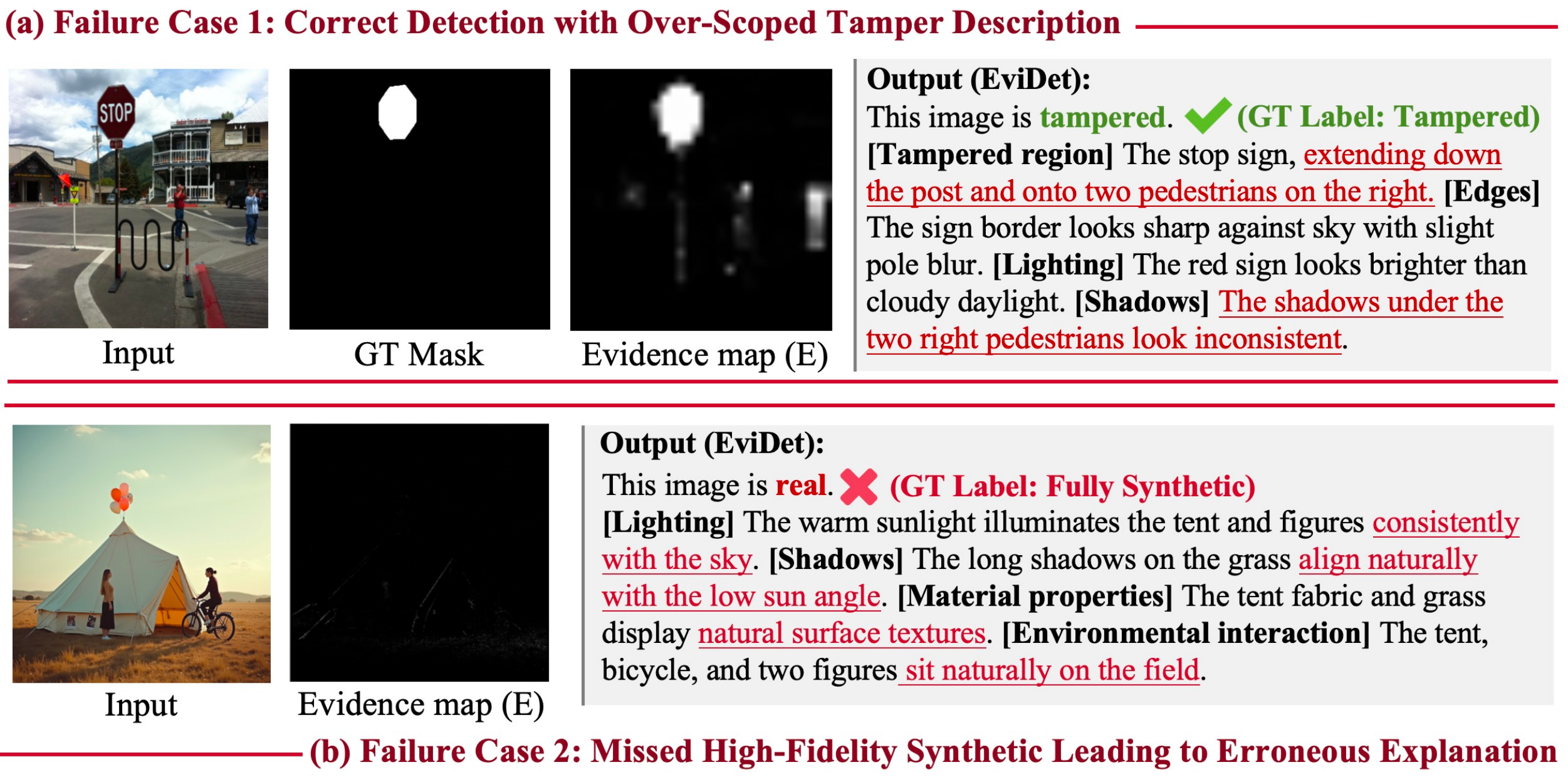}
\caption{Failure cases of EviDet. (a) Correct detection with an overly broad description of the edited region. (b) A visually plausible synthetic image misclassified as real, with weak evidence $E$ and an explanation aligned with the wrong label.}
\label{fig:failure-cases}
\end{figure*}

Due to space constraints, the main text presents only a subset of qualitative results. In this appendix, we provide additional examples to supplement the analysis. We first show several representative cases where EviDet provides more detailed reasoning, including the task-specific prompts and the detailed model outputs. We then present a comparative case that illustrates a characteristic trade-off of the contrastive decoder: while it preserves the core forensic evidence and final decision, it occasionally produces less comprehensive explanations compared to the ground-truth reference. These examples are intended to offer a more complete picture of the model behavior, including both its strengths and its practical limitations.

\noindent\textbf{Representative Cases.} Figure~\ref{fig:successful_cases} presents representative examples from all three classes. The class-dependent structure of $E$ is reflected in the generated explanations: near-null evidence supports the real assessment, distributed evidence grounds scene-wide synthetic artifacts, and localized evidence reflects the manipulated region in the tampered image.

\noindent\textbf{Trade-off Example of Contrastive Decoding.} As illustrated in Figure~\ref{fig:cd_tradeoff}, Contrastive Decoding (CD) and the ground-truth (GT) explanation both correctly classify the image as tampered and agree on the core evidence (lighting, edges, boundaries). However, CD omits the resolution cue and the descriptive motion detail present in GT, resulting in a less comprehensive and fluent explanation. This reflects a characteristic trade-off: the evidence map $E$ guides token selection toward strongly supported cues, occasionally suppressing details not directly anchored to the evidence. We will explore a diversity term or periodic context restoration to improve fluency while retaining guidance from the evidence map.

\noindent\textbf{Failure Cases.} In Figure~\ref{fig:failure-cases}(a), EviDet detects the edit and $E$ activates most strongly on the stop sign, yet the explanation extends the edit to the post and nearby pedestrians outside the annotated region. In (b), a visually plausible synthetic image is predicted as real with weak activation in $E$, and the explanation follows this error by describing the lighting and shadows as natural. These cases motivate more precise spatial descriptions and better detection of visually plausible synthetic images, which we leave for future work.

\section{Efficiency Analysis}

We isolate the inference cost of \textbf{EGE} by comparing decoding with and without CD. With CD enabled, images predicted as synthetic or tampered use the true and null evidence paths, while those predicted as real retain single-path decoding. We use three stratified SID-Set subsets (seeds 418, 2026, 3407) and keep all other settings fixed.

\begin{figure*}[htbp]
\centering
\includegraphics[width=0.9\textwidth]{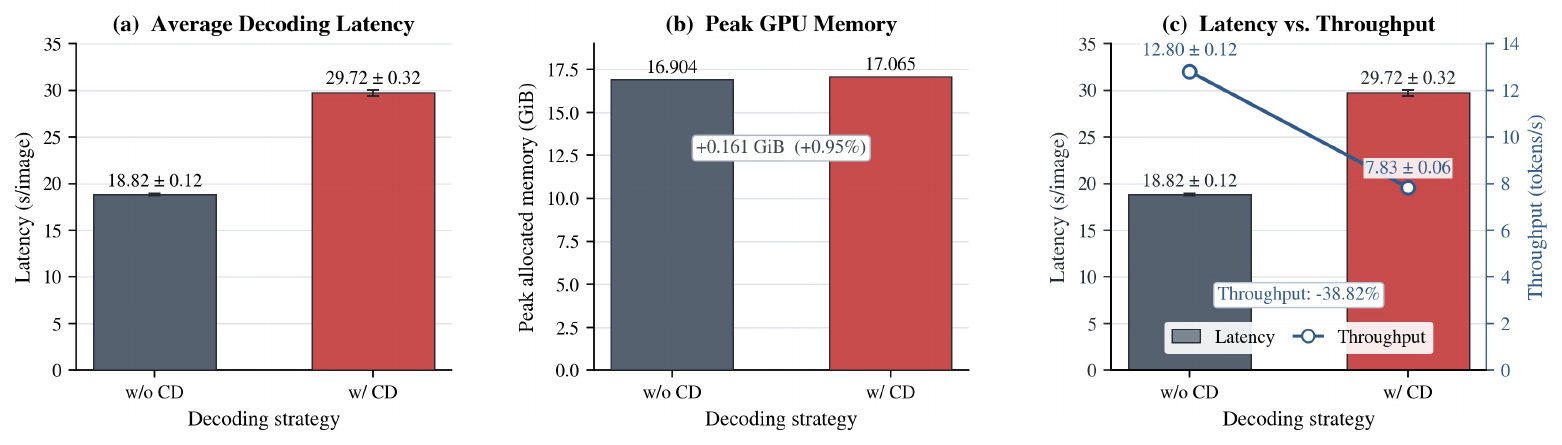}
\caption{Efficiency of EGE contrastive decoding on SID-Set. (a) Average latency. (b) Peak GPU memory. (c) Latency-throughput trade-off. Error bars show std over three seeds.}
\label{fig:cd-efficiency}
\end{figure*}

Table~\ref{tab:efficiency} and Figure~\ref{fig:cd-efficiency} report average latency, peak allocated GPU memory, output length, and decoding throughput. Latency measures complete explanation generation from cached EviDet features. The number of generated tokens per image captures output length, while tokens per second measures actual decoding throughput. 

The results show that CD increases latency from $18.82$ to $29.72$ s/image and reduces throughput from $12.80$ to $7.83$ tokens/s due to the additional null-evidence computation. While evaluating two evidence conditions, sharing model parameters limits the increase in peak memory to just $0.161$ GiB ($<1\%$) over the single-path. Together with the hallucination reduction reported in Table~5 of the main text, these results show that CD improves explanation reliability with minimal memory overhead, with additional decoding time as the main computational cost which we consider a worthwhile trade-off for more reliable offline forensic reporting.

\begin{table}[htbp]
\centering
\small
\setlength{\tabcolsep}{2.5pt}
\begin{tabular}{ccc}
\toprule
Metric & w/o CD ($\alpha=0$) & w/ CD ($\alpha=1$) \\
\midrule
Decoding \& Latency (s/img)$\downarrow$ & $18.82{\pm}0.12$ & $29.72{\pm}0.32$ \\
Peak memory (GiB)$\downarrow$ & $16.904$ & $17.065$ \\
Tokens/img & $240.93{\pm}0.75$ & $232.73{\pm}1.41$ \\
Tokens/s$\uparrow$ & $12.80{\pm}0.12$ & $7.83{\pm}0.06$ \\
\bottomrule
\end{tabular}
\caption{Inference cost of EGE contrastive decoding on SID-Set. Values are the mean and sample standard deviation over three independently stratified subsets. Memory denotes peak allocated GPU memory.}
\label{tab:efficiency}
\end{table}


\end{document}